\documentclass[sigconf,nonacm]{acmart}
\usepackage{multirow}
\usepackage[table]{xcolor}
\usepackage{makecell}
\usepackage[utf8]{inputenc} 
\usepackage[T1]{fontenc}    
\usepackage{hyperref}       
\usepackage{url}            
\usepackage{amsfonts}       
\usepackage{nicefrac}       
\usepackage{microtype}      
\usepackage{xcolor}         
\usepackage{subfigure}
\usepackage{amsmath}
\usepackage{amsthm} 
\usepackage{xurl}  
\usepackage{makecell}
\usepackage{enumitem}
\usepackage{pifont}
\usepackage{siunitx}         
\usepackage{threeparttable}
\usepackage{algorithm}
\usepackage{algpseudocode}
\algnewcommand\algorithmicinput{\textbf{Input:}}
\algnewcommand\algorithmicoutput{\textbf{Output:}}
\algnewcommand\Input{\item[\algorithmicinput]}
\algnewcommand\Output{\item[\algorithmicoutput]}

\newtheoremstyle{myremark}
  {2pt}   
  {2pt}   
  {}      
  {}      
  {\itshape} 
  {.}     
  {0.5em} 
  {}      

\theoremstyle{myremark}
\newtheorem{remark}{Remark}
\setlist{nosep}

\begin{document}

\title{IDDM: Identity-Decoupled Personalized Diffusion Models with a Tunable Privacy-Utility Trade-off}

\author{Linyan Dai}
\affiliation{%
  \institution{The Hong Kong Polytechnic University}
  \city{Hong Kong}
    \country{China}
}
\email{liann.dai@connect.polyu.hk}

\author{Xinwei Zhang}
\affiliation{
  \institution{The Hong Kong Polytechnic University}
  \city{Hong Kong}
    \country{China}
}
\email{xin-wei.zhang@connect.polyu.hk}

\author{Haoyang Li}
\affiliation{
  \institution{The Hong Kong Polytechnic University}
  \city{Hong Kong}
    \country{China}
}
\email{hao-yang9905.li@connect.polyu.hk}

\author{Qingqing Ye}
\affiliation{%
  \institution{The Hong Kong Polytechnic University}
  \city{Hong Kong}
    \country{China}
}
\email{qqing.ye@polyu.edu.hk}

\author{Haibo Hu}\authornote{Corresponding author.}
\affiliation{
  \institution{The Hong Kong Polytechnic University}
  \city{Hong Kong}
    \country{China}
}
\email{haibo.hu@polyu.edu.hk}

\begin{abstract}
Personalized text-to-image diffusion models (e.g., DreamBooth, LoRA) enable users to synthesize high-fidelity avatars from a few reference photos for social expression. However, once these generations are shared on social media platforms (e.g., Instagram, Facebook), they can be linked to the real user via face recognition systems, enabling identity tracking and profiling. Existing defenses mainly follow an \emph{anti-personalization} strategy that protects publicly released reference photos by disrupting model fine-tuning. While effective against unauthorized personalization, they do not address another practical setting in which personalization is authorized, but the resulting public outputs still leak identity information.

To address this problem, we introduce a new defense setting, termed \emph{model-side output immunization}, whose goal is to produce a personalized model that supports authorized personalization while reducing the identity linkability of public generations, with tunable control over the privacy-utility trade-off to accommodate diverse privacy needs. To this end, we propose \textbf{I}dentity-\textbf{D}ecoupled personalized \textbf{D}iffusion \textbf{M}odels \textbf{(IDDM)}, a model-side defense that integrates identity decoupling into the personalization pipeline. Concretely, IDDM follows an alternating procedure that interleaves short personalization updates with identity-decoupled data optimization, using a two-stage schedule to balance identity linkability suppression and generation utility. Extensive experiments across multiple datasets, diverse prompts, and state-of-the-art face recognition systems show that IDDM consistently reduces identity linkability while preserving high-quality personalized generation.

\end{abstract}





\maketitle

\section{Introduction}
\begin{figure}[t]
    \centering
\includegraphics[width=1\linewidth]{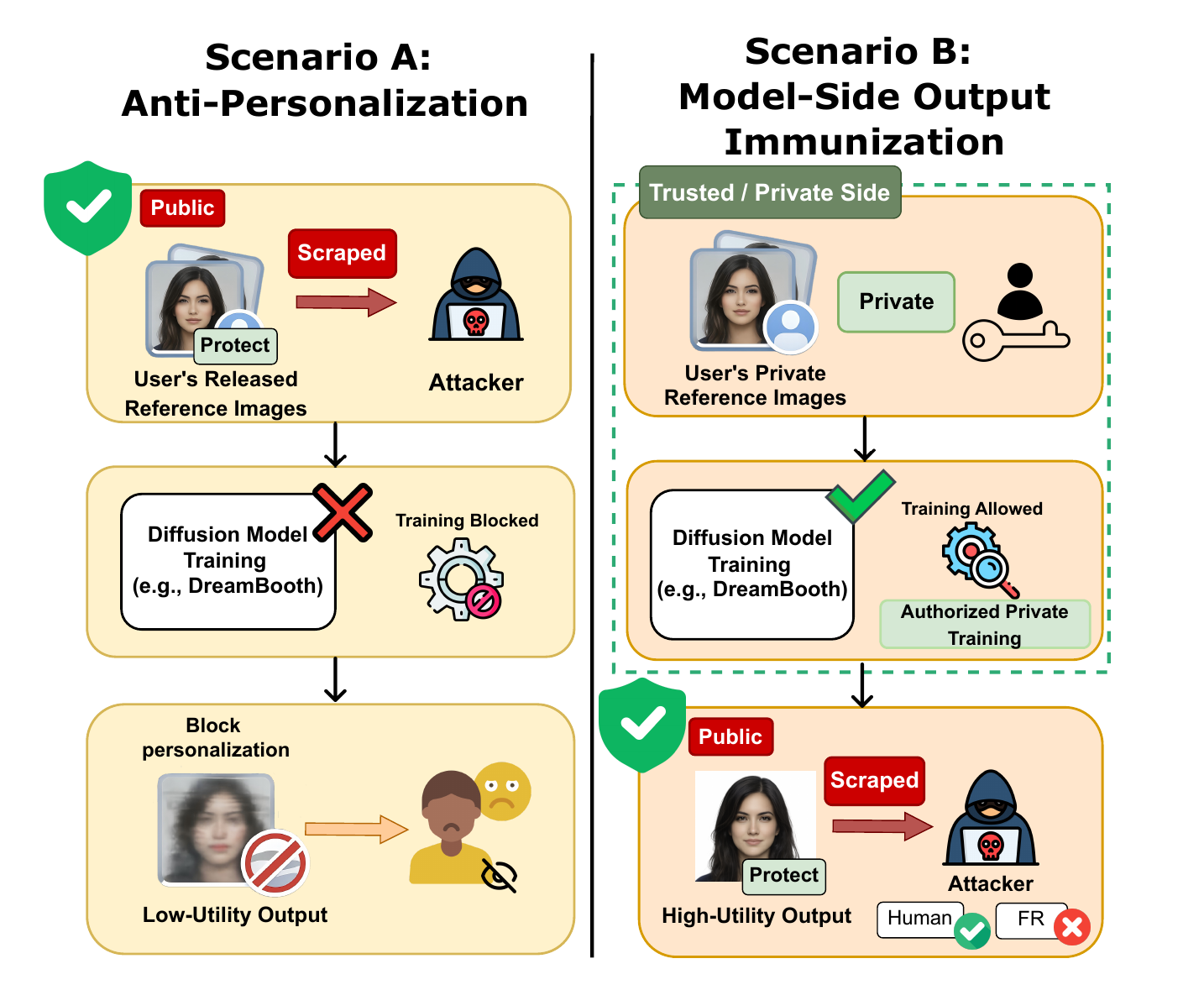}
    \caption{Comparison of protection paradigms under different threat models in personalized text-to-image generation: prior anti-personalization training (left) versus our model-side output immunization (right).}
    \label{fig:compare}
\end{figure}

Text-to-image diffusion models, such as Stable Diffusion \citep{compvis2022stable_diffusion_repo} and Nano Banana \citep{google_nanobanana_gemini_api}, have become a widely adopted paradigm for high-fidelity image synthesis \citep{rombach2022ldm, saharia2022imagen}. Beyond general-purpose generation, a particularly prominent application is subject-driven personalization, where users adapt a pretrained model to a specific identity or character from only a handful of reference images, enabling high-fidelity portraits and consistent characters for digital storytelling \citep{ruiz2023dreambooth, gal2022textualinversion}.
This approach is now common in creator communities. For example, Civitai reports about 3 million users and 12-13 million monthly unique visitors \citep{perez2023civitai}. The popularity of personalization is also clear from the number of trained diffusion models: a study reports that 868K+ LoRA models were trained in a single month on RunPod \citep{singh2025civitai_runpod}.

Despite this popularity, privacy concerns remain a major barrier, especially for face-centered personalization \citep{wang2024instantid,ruiz2023dreambooth,gal2022textualinversion,pang2025whiteboxmia,nguyen2019deepfakes_survey}. Personalized diffusion models are designed to reproduce a user's identity, so their outputs may retain recognizable biometric cues. Once publicly shared, such images can be scraped, redistributed, and linked back to the user by downstream face recognition systems. 
To mitigate this risk, prior work focuses on ``\textit{anti-personalization}'' by adding imperceptible perturbations to publicly released reference images to prevent unauthorized fine-tuning \citep{wang2024simac,vanle2023antidreambooth,myopia2025,xu2024caat,zheng2025antidiffusion,2025ppa}.

However, in practical personalization workflows, users often keep clean reference images private and share only the personalized outputs, especially in avatar and AI-headshot services. For example, during the Lensa boom, users uploaded a few selfies to a service without releasing them as reference images and publicly shared the generated portraits on social media \citep{Smith2022Lensa, Miltner2024Lensa}. In this setting, the privacy risk shifts from unauthorized training on released references to biometric cues preserved in the outputs. Motivated by this, we study a different defense setting, termed model-side output immunization. As illustrated in Figure \ref{fig:compare}, anti-personalization protects publicly released reference images to prevent downstream personalization at the data level (Scenario A), whereas our approach targets model-side output immunization under authorized personalization (Scenario B).

This shift in threat model motivates a new goal: \emph{identity-decoupled personalization}, where generated portraits remain visually useful and sufficiently faithful for creative applications, yet become harder to link back to the user by face recognition systems. Crucially, this goal requires a \emph{tunable} privacy-utility trade-off as different users may prefer different operating points, which is practical and yet underexplored in real-world creation. For instance, a user who publishes broadly or pseudonymously may prefer stronger protection against identity linkage, whereas a creator building a recurring avatar, branded persona, or story character may accept weaker privacy protection in exchange for higher identity consistency and fidelity. Therefore, a useful defense should allow authorized users to control how much identity signal is retained in their public outputs.

To achieve this goal, we propose \textbf{Identity-Decoupled personalized Diffusion Models (IDDM)}, a new model-side privacy mechanism for authorized personalization. IDDM integrates identity decoupling directly into the personalization pipeline, enabling users to adapt diffusion models to diverse creative needs with tunable privacy-utility trade-offs. Specifically, IDDM alternates short personalization updates with identity-decoupled data optimization on the user's private reference set, producing portrait outputs that remain visually useful while becoming less linkable to the user by downstream face recognition systems. To support tunable control, IDDM adopts a two-stage schedule in the identity-decoupled data optimization step, where the parameter $\rho$ adjusts the trade-off between privacy protection and generation utility: smaller $\rho$ emphasizes privacy, while larger $\rho$ better preserves utility. Furthermore, this model-side design eliminates the need for per-image post-processing, making it suitable for deployment by a trusted platform or service provider.

\begin{figure}[t]
    \centering
    \includegraphics[width=1\linewidth]{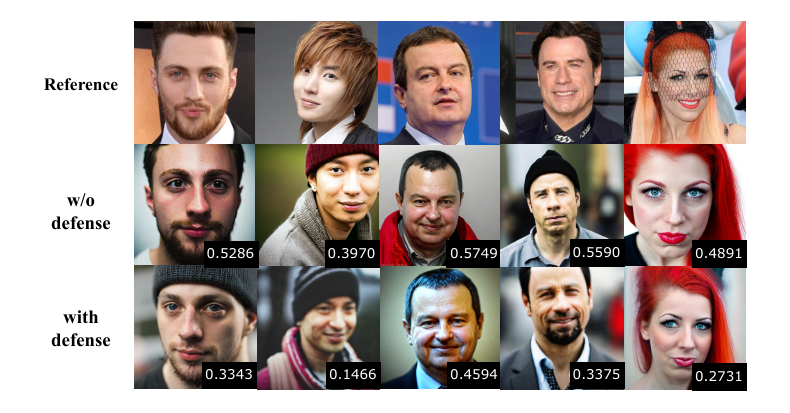}
    \caption{Visualization examples on VGGFace2 with \textit{``a dslr portrait of sks person"} under the same experimental setting. The white number below each generated image denotes the cosine similarity to the corresponding clean identity embedding (lower is less identity-related).}
    \label{example}
\end{figure}

We extensively evaluate IDDM across multiple datasets and state-of-the-art face recognition systems. The results demonstrate that IDDM achieves a favorable trade-off between portrait fidelity and identity protection. As shown in Figure~\ref{example}, compared with standard DreamBooth personalization (without defense), IDDM produces portraits that remain visually plausible while being harder to link back to the subject by downstream face recognition systems. In practice, IDDM can be integrated into vanilla DreamBooth/LoRA personalization by replacing the photo training procedure with an identity-decoupled version, while leaving the inference-time generation procedure unaltered.

In summary, our contributions are as follows:
\begin{itemize}[leftmargin=*]
    \item We formalize a new privacy setting for personalized diffusion, shifting protection from reference images to released outputs. Unlike prior defenses that disrupt personalization, our setting reduces the linkability of generated images to the user with a tunable privacy-utility trade-off.
    \item We propose Identity-Decoupled personalized Diffusion Models (IDDM), a model-side privacy mechanism that integrates identity decoupling into personalization pipeline, producing immunized outputs automatically without per-image inference-time post-processing.

    \item Extensive experiments show that IDDM consistently reduces identity linkability across multiple datasets, state-of-the-art recognizers, and diverse prompts, while maintaining high-quality personalized generation comparable to standard DreamBooth personalization. 
\end{itemize}
\section{Related Work}
\label{sec:related_work}

\subsection{Personalized Diffusion  Model}
Diffusion-based text-to-image models have become a dominant paradigm for high-fidelity image synthesis by learning to reverse a progressive noising process \citep{ho2020ddpm,rombach2022ldm}. 
Built on a pretrained diffusion backbone, personalization incorporates subject-specific information from only a few reference images, making the model render the same subject under diverse textual prompts. Some personalization methods keep the generator fixed and optimize the conditioning side. For instance, Text Inversion learns a new token embedding to represent a user-provided concept \citep{gal2022textualinversion}, and DeTeX improves disentanglement and controllability by learning more structured embeddings \citep{cai2024detex}. Another line fine-tunes parts of the denoiser so the model binds a unique identifier token to a specific subject. DreamBooth is a seminal method in this category and supports subject-driven generation under diverse prompts \citep{ruiz2023dreambooth}.
To reduce storage and training cost, many methods adopt parameter-efficient updates \citep{kumari2023customdiffusion,tewel2023perfusion,hu2022lora}.
For example, Custom Diffusion updates only a small subset of cross-attention parameters \citep{kumari2023customdiffusion}, and low-rank adaptation (LoRA) has become a common tool for efficient diffusion customization \citep{hu2022lora}. Recent work also explores lightweight alternatives that inject reference cues at inference time via auxiliary modules, improving usability while avoiding full fine-tuning \citep{ye2023ipadapter,li2024photomaker,wang2024instantid,zeng2024jedi}. Despite their effectiveness, personalized generators are inherently privacy-sensitive, as they aim to reproduce the fine-grained appearance of the subject from only a few reference images, which can encode identity-bearing traits \citep{ruiz2023dreambooth,vanle2023antidreambooth}. In this paper, we focus on the DreamBooth-style fine-tuning setting, which remains one of the most widely used personalization pipelines and is the primary target of our defense design and evaluation.

\subsection{Privacy Protection for Diffusion Models}
With personalized diffusion models becoming increasingly popular, privacy and misuse risks have arisen from pretraining and related downstream assignments. Some studies show that diffusion models may memorize and emit training examples, thereby facilitating data extraction and membership attacks \citep{carlini2023extracting,hu2023membership}. Accordingly, model-centric mitigations have been explored, including privacy-preserving training with differential privacy \citep{dockhorn2022dpdiffusion} and post-hoc removal of sensitive concepts via unlearning or model editing \citep{wu2025unlearning,gandikota2023erasing}.
In parallel, user-side defenses proactively protect images before uploading on social media by adding subtle noise into users' images to resist unauthorized malicious personalization, including AdvDM \citep{liang2023advdm}, Anti-DreamBooth \citep{vanle2023antidreambooth}, SimAC \citep{wang2024simac}, and CAAT \citep{xu2024caat}. MetaCloak further improves robustness by optimizing transformation-resilient perturbations for unauthorized subject-driven synthesis \citep{liu2024metacloak}. 
Beyond personalization, PhotoGuard \citep{salman2023photoguard} immunizes images against diffusion-based editing, whereas Anti-Diffusion \citep{zheng2025antidiffusion} targets broader misuse spanning both tuning and editing modifications.
Overall, existing user-side protections are designed for a deny-adaptation objective (i.e., to block downstream personalization or editing), thus sacrificing the utility of authorized users who still require high-fidelity personalization.
\section{Preliminary and Threat Model}
\label{p1}

\subsection{Diffusion Models}
Diffusion models are probabilistic generative models that can create high-quality images from Gaussian noise by reversing a forward noising Markov chain. Given input image $x_0 \sim q(x)$, the forward process adds increasing noise to inputs by the noise scheduler $\{\beta_t: \beta_t \in (0,1)\}_{t=1}^T$ to produce a series of noisy variables over T steps: $\{x_1,x_2,\dots, x_T\}$, and $x_t$ can be formulated at timestep $t$ as follows:
\begin{equation}
    x_t=\sqrt{\bar{\alpha_t}}x_0 + \sqrt {1-\bar{\alpha_t}}\epsilon, 
\end{equation}
where $\alpha_t=1-\beta_t$, $\bar{\alpha_t}=\prod_{s=1}^t \alpha_s$ and $\epsilon \sim \mathcal{N}(0, \mathbf{I})$.

In the backward process, the model learns to denoise the noisy sample $x_{t+1}$ from the previous noisy sample $x_t$ by training a noise prediction neural network $\epsilon_\theta(x_{t+1},t)$ to obtain the desired injected noise $\epsilon$. Particularly, the denoising process can be formulated as an optimization objective of minimizing the $\ell_2$ distance between the predicted noise and true noise:
\begin{align}
    &\mathcal{L}_{\text{uncond}}(\theta, x_0) = \mathbb{E}_{x_0, t, \epsilon \sim \mathcal{N}(0, \mathbf{I})} \left\| \epsilon - \epsilon_\theta(x_{t+1}, t) \right\|_2^2, \\
    &\mathcal{L}_{\text{cond}}(\theta, x_0) = \mathbb{E}_{x_0, t, c, \epsilon \sim \mathcal{N}(0, \mathbf{I})} \left\| \epsilon - \epsilon_\theta(x_{t+1}, t, c) \right\|_2^2,\label{f2}
\end{align}
where $c$ denotes the conditional prompt directing the generation.

\subsection{Subject-Driven Personalization}

Subject-driven personalization adapts a pretrained text-to-image diffusion model to a specific subject from only a few reference images. DreamBooth \citep{ruiz2023dreambooth} is a representative formulation, which fine-tunes the model using subject images paired with identifier prompts as depicted in (\ref{f2}), together with a prior-preservation term to retain class diversity. Its objective can be written as
\begin{align}
    \mathcal{L}_{db}(\theta, x_0) = \mathbb{E}_{x_0, t, t^\prime}  \left\| \epsilon - \epsilon_\theta(x_{t+1}, t, c) \right\|_2^2 \notag\\
    +\lambda \left\| \epsilon^\prime - \epsilon_\theta(x^\prime_{t^\prime+1}, t^\prime, c_{pr}) \right\|_2^2,\label{prior}
\end{align}
where $\epsilon$ and $\epsilon^\prime $ denote noise samples from $\mathcal{N}(0, \mathbf{I})$, $x^\prime _{t^\prime +1}$ is the noisy state originated from the class sample $x^\prime $ over $t^\prime +1$ timesteps, and $c_{pr}$ is prior prompt. In practice, personalization can be implemented either by full-model fine-tuning (e.g., DreamBooth \citep{ruiz2023dreambooth}) or by parameter-efficient adaptation (e.g., LoRA \citep{hu2022lora}).


\subsection{Threat Model}
\label{sec:threat_model}
We consider a setting in which a user provides a small set of private photos $\mathcal{X}$ to an authorized personalization pipeline in order to obtain a personalized text-to-image diffusion model. The private photos $\mathcal{X}$ are not publicly released; instead, the public exposure arises from the generated images $\mathcal{X}_{gen}$ produced by the resulting personalized model. Accordingly, the privacy risk in our setting stems from the public dissemination of $\mathcal{X}_{gen}$ rather than leakage of the private training photos or the personalized model parameters.
\noindent\textbf{Adversary.}
The adversary can scrape released images $\mathcal{X}_{gen}$ from the web and run a standard face-analysis pipeline, including face detection, alignment, feature extraction, and matching, to link them to the user’s real-world identity. We assume the adversary may query black-box state-of-the-art Face Recognition Systems (FRS), but has no access to the user’s private photos $\mathcal{X}$, personalization prompts, or the personalized model parameters and gradients.

\noindent\textbf{Defender.}
The defender is either the user or a trusted personalization service provider. Given access to the user's private photos $\mathcal{X}$ and the authorized personalization pipeline, the defender is allowed to modify the personalization procedure to produce an \emph{identity-decoupled personalized model}. The goal is to preserve the utility of personalized generation while reducing the linkability of its released outputs to the user by downstream FRS. Depending on the user's privacy preferences or deployment requirements, the defender can further adjust the defense strength to realize a desired trade-off between generation utility and identity protection.

Importantly, the defense is \emph{model-side}: once the identity-decoupled personalized model has been obtained, its outputs are protected by default under normal use, without requiring per-image inference-time post-processing before release. To optimize the defense, the defender may use a set of surrogate face-recognition encoders, but does not know the exact recognizer deployed by the adversary.

\section{Our Method: IDDM}
\label{sec:method}
\begin{figure*}[t]
    \centering
    \includegraphics[width=1\linewidth]{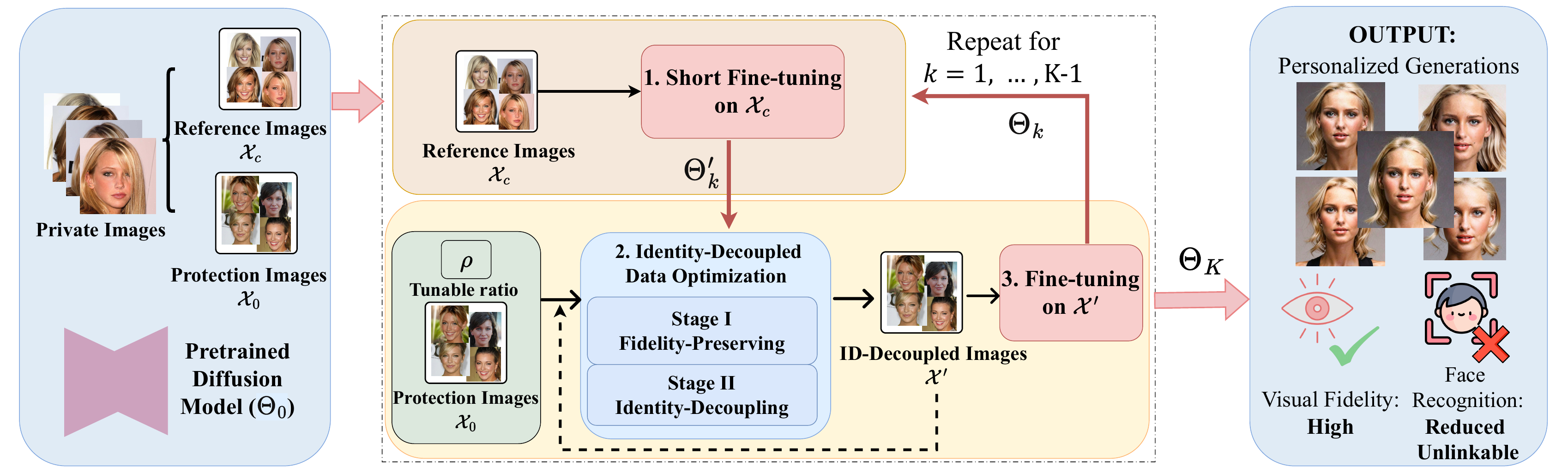}
    \caption{The framework of our proposed IDDM. Each iteration alternates between a short personalization update on the fixed reference set $\mathcal{X}_c$ (Step~1) and an identity-decoupled data update on the protection set $\mathcal{X}_0$, followed by fine-tuning on the updated set $\mathcal{X}'$ (Steps~2--3). In Step~2, the tunable stage split ratio $\rho$ determines the relative allocation of optimization between Stage~I and Stage~II, allowing IDDM to operate at different privacy-utility trade-off points.}
    \label{fig:IDDM}
\end{figure*}
\subsection{Overview}
The key idea of our method is to embed identity decoupling into \emph{authorized personalization} by optimizing the user’s \textit{private} training photos within a small bounded budget, so that the resulting personalized diffusion model maintains generation utility while reducing identity linkability in its outputs.

Figure~\ref{fig:IDDM} summarizes the overall procedure.
IDDM follows an alternating loop over $K$ iterations that interleaves personalization updates with identity-decoupled data optimization.
At iteration $k$, we first run a short fine-tuning on a fixed reference set $\mathcal{X}_c$ to obtain a temporary model state $\Theta'_k$ that reflects the current personalization stage (Step~1). We then freeze $\Theta'_k$ and update the protection set $\mathcal{X}_0$ under an $\ell_\infty$ constraint to obtain an identity-decoupled training set $\mathcal{X}'$ (Step~2).  Conditioning this step on $\Theta'_k$ is essential, since identity leakage arises from the evolving personalization trajectory rather than from a fixed pretrained model. Finally, we fine-tune on $\mathcal{X}'$ to obtain the updated personalized model $\Theta_k$ (Step~3), thereby incorporating the current identity-decoupling effect into the model before the next iteration.


Section~\ref{sec:alt_train} details this alternating personalization-aware training loop, and Section~\ref{sec:id_decoupled_data} specifies the two-stage optimization in Step~2. This two-stage design explicitly balances fidelity preservation and identity-linkability suppression, yielding a more stable optimization process and a better utility-privacy trade-off than joint optimization, as shown in Table \ref{reason} in Appendix~A.2. 

\begin{remark}
IDDM differs from prior anti-personalization methods, despite using projected gradient updates to optimize training images in Step~2. Existing anti-personalization defenses apply protective perturbations to publicly released photos in order to disrupt unauthorized fine-tuning, whereas IDDM optimizes the user's \emph{private} training photos only as internal variables within an authorized personalization pipeline. The optimized identity-decoupled image set $\mathcal{X}'$ is not published or exposed to the attacker, and it is used solely to shape the resulting personalized model toward reduced identity linkability. Therefore, as a model-side defense, IDDM does not rely on the persistence of released perturbations and is not directly affected by purification or denoising of public images at deployment time.
\end{remark}

\subsection{Alternating Personalization-Aware Training}
\label{sec:alt_train}
To preserve utility, our defense must be compatible with standard personalization (e.g., DreamBooth/LoRA), rather than blocking fine-tuning.
However, personalization continuously updates model parameters and can (re-)introduce identity cues as training proceeds, which may weaken protections optimized for an earlier model snapshot.
Therefore, we adopt an alternating training scheme that updates the protection images using a model aligned with the current personalization stage, so that the protection persists after subsequent fine-tuning.

Given a user’s private photo set $\mathcal{X}=\{x_i\}_{i=1}^{N}$, we split it into two disjoint subsets
$\mathcal{X}=\mathcal{X}_0 \cup \mathcal{X}_c$ with $\mathcal{X}_0 \cap \mathcal{X}_c=\varnothing$.
Here, $\mathcal{X}_c$ is a \emph{fixed reference set} used for (i) short fine-tuning to approximate the current personalization state and (ii) identity prototype construction in our face-recognition space, while $\mathcal{X}_0$ is the \emph{protection set} whose samples will be updated under a bounded budget.

IDDM proceeds in $K$ iterations, starting from the pretrained diffusion model $M_{\Theta_0}$.
At iteration $k$, we maintain the current model parameters $\Theta_k$ and perform three steps (Figure~\ref{fig:IDDM}):

\noindent\textbf{Step 1: Short fine-tuning on $\mathcal{X}_c$.}
We initialize a temporary model from the current parameters and run a short personalization update on the fixed reference set:
\begin{equation}
\Theta'_k \leftarrow \operatorname{FT}(\Theta_{k-1};\mathcal{X}_c),
\end{equation}
where $\operatorname{FT}(\cdot)$ denotes a few-step DreamBooth/LoRA fine-tuning.
The resulting $\Theta'_k$ approximates the current fine-tuning state and provides an up-to-date model for guiding the subsequent data update.

\noindent\textbf{Step 2: Identity-decoupled data update on $\mathcal{X}_0$.}
We then \emph{freeze} the temporary model $M_{\Theta'_k}$ and update the protection set under the bounded budget (Section~\ref{sec:id_decoupled_data}):
\begin{equation}
\mathcal{X}' \leftarrow \operatorname{Update}(\mathcal{X}_0; M_{\Theta'_k}),
\end{equation}
where $\operatorname{Update}(\cdot)$ is a two-stage procedure (Stage~I fidelity-preserving, followed by Stage~II identity-decoupling) producing ID-decoupled training photos $\mathcal{X}'$. We refer to $\mathcal{X}'$ as an ID-decoupled dataset since fine-tuning on $\mathcal{X}'$ gets a personalized model whose outputs retain visual utility while being less linkable to the user in the downstream face-recognition space.

\noindent\textbf{Step 3: Fine-tuning on $\mathcal{X}'$.}
Finally, we continue personalization using the updated photos to obtain the next-iteration model:
\begin{equation}
\Theta_{k} \leftarrow \operatorname{FT}(\Theta'_k;\mathcal{X}').
\end{equation}
This step ensures that the model used in the next iteration reflects training on the latest protected set, preventing the data update from optimizing against a stale personalization state.

By repeatedly refreshing $\Theta'_k$ and updating $\mathcal{X}_0$ accordingly, IDDM couples training-data adaptation with the evolving personalization process. This prevents optimization against a stale model state and helps the identity-decoupling effect persist throughout subsequent fine-tuning. Algorithm \ref{outer_k} in Appendix~B summarizes this alternating optimization procedure for IDDM.

\subsection{Identity-Decoupled Data Optimization}
\label{sec:id_decoupled_data}

At each iteration $k$, we freeze the temporary model $M_{\Theta'_k}$ and update the protection set $\mathcal{X}_0$ under an $\ell_\infty$ budget $\|\delta\|_\infty\le \eta$, yielding a identity-decoupled set $\mathcal{X}'$ for authorized personalization.
For each $x\in\mathcal{X}_0$, we denote its identity-decoupled version as $x'=\Pi_{[0,1]}(x+\delta)$. 
As shown in Table \ref{reason} in Appendix A.2, optimizing identity suppression from the start (i.e., jointly optimization) often distorts facial structure and causes a large fidelity drop, leading to unstable or conflicting gradients. Therefore, we use a two-stage schedule to balance utility and identity privacy: stage~I regularizes the identity-decoupled training set with a denoising-consistency objective so that they remain compatible with diffusion training, and stage~II reduces identity linkability in the face-recognition feature space based on denoised predictions.

\noindent\textbf{Stage I: Fidelity-Preserving Optimization.}
For latent diffusion models, we first encode $x'$ into a latent $z_0$ via the VAE encoder.
Given a timestep $t$ and Gaussian noise $\epsilon\sim\mathcal{N}(0,\mathbf{I})$, we form
$z_t=\sqrt{\bar{\alpha}_t}\,z_0+\sqrt{1-\bar{\alpha}_t}\,\epsilon$,
where $\bar{\alpha}_t$ denotes the cumulative noise coefficient.
Let $\epsilon_{\Theta'_k}(\cdot)$ denote the denoising network of the frozen temporary model conditioned on prompt $c$.
The reconstruction (denoising) objective is:
\begin{align}
\label{eq:rec}
\mathcal{L}_{\mathrm{rec}}(\Theta'_k, x')
= \mathbb{E}_{t,\epsilon}\Big[ \big\| \epsilon - \epsilon_{\Theta'_k}(z_t,t,c) \big\|_2^2 \Big].
\end{align}
This objective regularizes bounded data adaptation to ensure denoising consistency, so that the resulting training photos remain compatible with diffusion fine-tuning, helping mitigate fidelity degradation after identity decoupling. We obtain $\delta$ by optimizing:
\begin{align}
\arg\min_{\delta}\ \mathcal{L}_{\mathrm{rec}}(\Theta'_k, x')\quad
\text{s.t.}\ \|\delta\|_\infty \le \eta.
\end{align}

\noindent\textbf{Stage II: Identity-Decoupling Optimization.}
Let $\{f_m\}_{m=1}^{M}$ be a set of pretrained face-recognition (FR) encoders (e.g., MobileFace~\citep{chen2018mobilefacenets} and FaceNet~\citep{schroff2015facenet}), and let $\mathcal{A}(\cdot)$ denote a face pre-processing function that detects, crops, and aligns the face region before feeding it to an FR encoder. We form an identity prototype for each encoder using the aligned reference set $\mathcal{X}_c=\{x_j\}_{j=1}^{N_c}$:
\begin{equation}
e_m = \frac{1}{N_c}\sum_{j=1}^{N_c} f_m\!\big(\mathcal{A}(x_j)\big),
\end{equation}
where $f_m(\cdot)$ outputs an $\ell_2$-normalized embedding and $e_m$ is normalized to unit length.

During optimization, given an adapted training image $x'$ obtained after stage I, we run a denoising step using the frozen temporary model $M_{\Theta'_k}$ and obtain the corresponding clean prediction $\hat{x}_0$
(e.g., by decoding the predicted $\hat{z}_0$ through the VAE decoder).
For each face recognizer $f_m$, we measure how easily the current sample can be linked to the user by the cosine similarity between its embedding and the identity prototype:
\begin{equation}
s_m(x') = \cos\!\Big(f_m\big(\mathcal{A}(\hat{x}_0)\big),\, e_m\Big)\in[-1,1].
\end{equation}
A larger $s_m(x')$ indicates higher identity linkability under the $m$-th recognizer.

We then aggregate the ensemble responses with adaptive weights so that the most vulnerable recognizer (the one giving the largest similarity) has a larger influence on the objective, while keeping the aggregation differentiable
\begin{align}
w_m(x') &= \frac{\exp\!\big(s_m(x')/\tau\big)}{\sum_{j=1}^{M}\exp\!\big(s_j(x')/\tau\big)}, \notag\\
S(x') &= \sum_{m=1}^{M} w_m(x')\, s_m(x').\label{eq:weight}
\end{align}
Here $\tau>0$ controls how peaked the weights are, where a smaller $\tau$ makes $S(x')$ focus more on the largest $s_m(x')$.
Although $w_m$ increases with $s_m$, we minimize $S(x')$, so this weighting concentrates gradients on the most identity-sensitive recognizer and drives its similarity down.

We define the identity-decoupling objective as
\begin{equation}
\mathcal{L}_{\mathrm{id}}(\Theta'_k, x') = 1 + S(x').
\label{eq:id}
\end{equation}
Since $S(x')\in[-1,1]$, $(1+S(x'))\in[0,2]$ provides a stable, bounded objective.
Minimizing $\mathcal{L}_{\mathrm{id}}$ encourages low similarity to the identity prototypes across the recognizer ensemble.
We optimize the bounded adaptation by solving
\begin{align}
\arg\min_{\delta}\ \mathcal{L}_{\mathrm{id}}(\Theta'_k, x')\quad
\text{s.t.}\ \|\delta\|_\infty \le \eta,
\end{align}
where $x'=\Pi_{[0,1]}(x+\delta)$ for each $x\in\mathcal{X}_0$.

\noindent\textbf{Two-stage schedule.}
To balance utility preservation and identity suppression, we run projected gradient descent (PGD) on $\delta$ for $T$ steps.
We use a split ratio $\rho\in(0,1)$ to allocate the optimization budget between the two stages, where $\rho T$ steps are used for Stage~I (fidelity-preserving) and $(1-\rho)T$ steps for Stage~II (identity-decoupling).
For the first $T_{\mathrm{rec}}=\lfloor \rho T\rfloor$ steps, we optimize $\delta$ using $\mathcal{L}_{\mathrm{rec}}$ in Eq.~(\ref{eq:rec});
for the remaining steps, we switch to $\mathcal{L}_{\mathrm{id}}$ in Eq.~(\ref{eq:id}).
Each update is followed by the projection $\Pi_{\eta}(\cdot)$ to enforce $\|\delta\|_\infty\le \eta$.
The resulting identity-decoupled training set is denoted as $\mathcal{X}'=\{x'_i\}_{i=1}^{N_0}$, which is then used for authorized personalization in Step~3. 
Here, the FR objective is used only as a differentiable proxy for identity linkability under frozen recognizers, rather than as a target for standalone adversarial evasion. This staged update avoids directly optimizing two objectives in a single loss.
In face-centered personalization, reducing FR similarity often requires modifying identity-discriminative facial cues, while these cues are also vital for generation fidelity and subject consistency.
As a result, jointly optimizing the two objectives can suffer from gradient tug-of-war and unstable trade-offs (see Table \ref{reason} in Appendix~A.2).

\begin{table*}[t]
\centering
\caption{Defense performance comparison with anti-personalization baselines on CelebA-HQ and VGGFace2 datasets. IDDM is reported under three values of the preset step-allocation ratio $\rho$, which controls the trade-off between defense strength and generation utility. FSR measures output utility; ISM and ADA denote identity-related metrics; FID, SER, and BRQ denote the visual quality of generated images. }
\label{tab:defense_results}
\setlength{\tabcolsep}{2.6pt}
\renewcommand{\arraystretch}{1.10}
\small
\begin{threeparttable}
\begin{tabular*}{\textwidth}{@{\extracolsep{\fill}}llcccccccccccc}
\toprule
\multirow{2}{*}{\textbf{Dataset}} & \multirow{2}{*}{\textbf{Method}} &
\multicolumn{6}{c}{\textit{``a photo of sks person''}} &
\multicolumn{6}{c}{\textit{``a dslr portrait of sks person''}} \\
\cmidrule(lr){3-8} \cmidrule(lr){9-14}
& &
FSR$\uparrow$ & ISM$\downarrow$ & ADA$\downarrow$ & FID$\downarrow$ & SER$\uparrow$ & BRQ$\downarrow$ &
FSR$\uparrow$ & ISM$\downarrow$ & ADA$\downarrow$ & FID$\downarrow$ & SER$\uparrow$ & BRQ$\downarrow$ \\
\midrule

\multirow{7}{*}{CelebA-HQ}
& No Defense
& 0.57 & 0.69 & 0.66 & 158.09 & 0.76 & 10.65
& 0.50 & 0.48 & 0.44 & 214.35 & 0.74 & 4.88 \\
& Anti-DB
& 0.50 & 0.38 & 0.34 & 314.67 & 0.52 & 40.21
& 0.45 & 0.28 & 0.24 & 341.23 & 0.59 & 36.06 \\
& SimAC
& 0.30 & 0.21 & 0.17 & 395.81 & 0.27 & 37.34
& 0.15 & \textbf{0.03} & \textbf{0.03} & 458.58 & 0.22 & 42.23 \\
& Anti-Diff
& 0.33 & \textbf{0.13} & \textbf{0.11} & 383.80 & 0.26 & 41.96
& 0.32 & 0.16 & 0.13 & 417.32 & 0.38 & 37.70 \\
\cmidrule{2-14}
& IDDM ($\rho=0.1$)
& \textbf{0.65} & 0.43 & 0.38 & 213.37 & 0.74 & \textbf{16.81}
& 0.48 & 0.29 & 0.25 & 237.54 & 0.75 & \textbf{11.71} \\
& IDDM ($\rho=0.3$)
& 0.62 & 0.51 & 0.47 & 172.80 & \textbf{0.75} & 21.31
& 0.52 & 0.36 & 0.33 & 208.40 & 0.75 & 12.56 \\
& IDDM ($\rho=0.5$)
& 0.62 & 0.53 & 0.49 & \textbf{171.77} & \textbf{0.75} & 20.87
& \textbf{0.56} & 0.40 & 0.37 & \textbf{203.89} & \textbf{0.77} & 13.65 \\

\midrule

\multirow{7}{*}{VGGFace2}
& No Defense
& 0.75 & 0.69 & 0.65 & 186.50 & 0.73 & 22.71
& 0.47 & 0.48 & 0.44 & 249.59 & 0.69 & 8.07 \\
& Anti-DB
& 0.62 & 0.44 & 0.40 & 324.03 & 0.48 & 37.36
& 0.49 & 0.30 & 0.26 & 351.49 & 0.54 & 34.27 \\
& SimAC
& 0.36 & 0.24 & 0.20 & 375.84 & 0.27 & 38.74
& 0.15 & \textbf{0.02} & \textbf{0.01} & 431.06 & 0.17 & 41.59 \\
& Anti-Diff
& 0.44 & \textbf{0.22} & \textbf{0.18} & 380.42 & 0.34 & 41.54
& 0.36 & 0.18 & 0.15 & 393.11 & 0.49 & 37.13 \\
\cmidrule{2-14}
& IDDM ($\rho=0.1$)
& \textbf{0.81} & 0.49 & 0.45 & 216.68 & \textbf{0.73} & \textbf{21.68}
& 0.57 & 0.34 & 0.30 & 260.18 & 0.71 & 14.17 \\
& IDDM ($\rho=0.3$)
& \textbf{0.81} & 0.55 & 0.51 & \textbf{185.65} & \textbf{0.73} & 28.31
& \textbf{0.59} & 0.42 & 0.38 & \textbf{228.48} & \textbf{0.75} & \textbf{14.09} \\
& IDDM ($\rho=0.5$)
& 0.77 & 0.56 & 0.51 & 186.15 & \textbf{0.73} & 30.12
& 0.57 & 0.46 & 0.39 & 234.04 & 0.73 & 14.93 \\
\bottomrule
\end{tabular*}

\end{threeparttable}
\end{table*}

\begin{table*}[t]
\centering
\caption{Defense performance of IDDM under four additional prompts with different values of the tunable step-allocation ratio $\rho$ on CelebA-HQ. ``No Defense'' denotes the standard DreamBooth personalization without defense.}
\label{tab:extra_prompts_rho}
\small
\setlength{\tabcolsep}{4.5pt}
\renewcommand{\arraystretch}{1.15}
\begin{threeparttable}
\begin{tabular*}{\textwidth}{@{\extracolsep{\fill}}l
  *{2}{S[table-format=1.2] S[table-format=1.2] S[table-format=1.2] S[table-format=3.2] S[table-format=1.2] S[table-format=2.2]}}
\toprule
\multirow{2}{*}{\textbf{Method}} &
\multicolumn{6}{c}{\textit{``a photo of sks person wearing a hat''}} &
\multicolumn{6}{c}{\textit{``a photo of sks person with glasses''}} \\
\cmidrule(lr){2-7}\cmidrule(lr){8-13}
& {FSR$\uparrow$} & {ISM$\downarrow$} & {ADA$\downarrow$} & {FID$\downarrow$} & {SER$\uparrow$} & {BRQ$\downarrow$}
& {FSR$\uparrow$} & {ISM$\downarrow$} & {ADA$\downarrow$} & {FID$\downarrow$} & {SER$\uparrow$} & {BRQ$\downarrow$} \\
\midrule
No Defense
& 0.73 & 0.54 & 0.50 & 273.25 & 0.70 & 8.31
& 0.33 & 0.55 & 0.50 & 208.03 & 0.75 & 13.89 \\

IDDM ($\rho=0.1$)
& 0.70 & \textbf{0.34} & \textbf{0.29} & 275.65 & 0.70 & \textbf{12.31}
& 0.29 & \textbf{0.35} & \textbf{0.30} & 235.79 & 0.75 & \textbf{15.43} \\
IDDM ($\rho=0.3$)
& \textbf{0.73} & 0.44 & 0.41 & 272.79 & \textbf{0.72} & 18.18
& \textbf{0.30} & 0.45 & 0.41 & 194.94 & \textbf{0.77} & 19.72 \\
IDDM ($\rho=0.5$)
& 0.70 & 0.47 & 0.44 & \textbf{260.65} & \textbf{0.72} & 20.85
& 0.26 & 0.49 & 0.44 & \textbf{192.21} & 0.76 & 22.39 \\
\specialrule{\heavyrulewidth}{0pt}{0pt}
\multirow{2}{*}{\textbf{Method}} &
\multicolumn{6}{c}{\textit{``a photo of sks person with red hair''}} &
\multicolumn{6}{c}{\textit{``a photo of sks person with big eyes''}} \\
\cmidrule(lr){2-7}\cmidrule(lr){8-13}
& {FSR$\uparrow$} & {ISM$\downarrow$} & {ADA$\downarrow$} & {FID$\downarrow$} & {SER$\uparrow$} & {BRQ$\downarrow$}
& {FSR$\uparrow$} & {ISM$\downarrow$} & {ADA$\downarrow$} & {FID$\downarrow$} & {SER$\uparrow$} & {BRQ$\downarrow$} \\
\midrule
No Defense
& 0.65 & 0.58 & 0.54 & 189.17 & 0.72 & 10.04
& 0.08 & 0.48 & 0.45 & 267.91 & 0.59 & 14.10 \\

IDDM ($\rho=0.1$)
& 0.54 & \textbf{0.39} & \textbf{0.35} & 195.30 & 0.71 & \textbf{15.28}
& \textbf{0.11} & \textbf{0.29} & \textbf{0.24} & 280.63 & 0.59 & \textbf{14.99} \\
IDDM ($\rho=0.3$)
& \textbf{0.63} & 0.49 & 0.45 & 175.01 & 0.72 & 23.25
& 0.10 & 0.42 & 0.37 & 250.88 & 0.60 & 15.25 \\
IDDM ($\rho=0.5$)
& 0.59 & 0.53 & 0.49 & \textbf{171.66} & \textbf{0.73} & 24.61
& 0.10 & 0.49 & 0.45 & \textbf{229.81} & \textbf{0.64} & 16.85 \\
\bottomrule
\end{tabular*}
\end{threeparttable}
\end{table*}

\begin{table*}[t]
\centering
\caption{Defense performance comparison with post-processing baselines on CelebA-HQ. IDDM is reported under $\rho=0.1$.}
\label{tab:postprocess_main_metrics}
\vspace{-4pt}
\small
\setlength{\tabcolsep}{4.5pt}
\renewcommand{\arraystretch}{1.15}
\begin{tabular*}{\textwidth}{@{\extracolsep{\fill}}lcccccccccc@{}}
\toprule
\multirow{2}{*}{\textbf{Method}}
& \multicolumn{5}{c}{\textit{`a photo of sks person''}} 
& \multicolumn{5}{c}{\textit{`a dslr portrait of sks person''}} \\
\cmidrule(lr){2-6} \cmidrule(lr){7-11}
& ISM$\downarrow$ & ADA$\downarrow$ & FID$\downarrow$ & SER$\uparrow$ & BRQ$\downarrow$
& ISM$\downarrow$ & ADA$\downarrow$ & FID$\downarrow$ & SER$\uparrow$ & BRQ$\downarrow$ \\
\midrule
No Defense
& 0.69 & 0.66 & 158.09 & 0.76 & 10.65
& 0.48 & 0.44 & 214.35 & 0.74 & 4.88 \\

AdvDM
& \textbf{0.38} & \textbf{0.35} & 228.65 & 0.74 & \underline{14.73}
& 0.42 & 0.39 & 254.70 & 0.72 & \underline{14.82} \\

LowKey
& 0.64 & 0.62 & 214.04 & 0.74 & \textbf{8.23}
& 0.44 & 0.42 & 255.95 & 0.74 & 17.14 \\

GIFT
& \underline{0.40} & \underline{0.36} & \textbf{170.79} & \textbf{0.75} & 32.71
& \underline{0.29} & \underline{0.26} & \textbf{216.42} & 0.74 & 39.62 \\

IDDM
& 0.43 & 0.38 & \underline{213.37} & 0.74 & 16.81
& \textbf{0.28} & \textbf{0.25} & \underline{237.54} & \textbf{0.75} & \textbf{11.71} \\
\bottomrule
\end{tabular*}
\vspace{-4pt}
\end{table*}

\begin{table*}[t]
\centering
\caption{True Accept Rate (TAR, \%) at FAR=1e-4 across seven face recognition models on CelebA-HQ. }
\label{tab:new_evaluation}
\setlength{\tabcolsep}{3pt}
\renewcommand{\arraystretch}{1.15}
\begin{tabular*}{\textwidth}{@{\extracolsep{\fill}}l*{16}{c}@{}}
\toprule
\multirow{2}{*}{\textbf{Method}}
& \multicolumn{8}{c}{\textit{``a photo of sks person''}}
& \multicolumn{8}{c}{\textit{``a dslr portrait of sks person''}} \\
\cmidrule(lr){2-9} \cmidrule(lr){10-17}
& Ada.$\downarrow$ & QMag.$\downarrow$ & Elas.$\downarrow$ & R18$\downarrow$
& R50$\downarrow$ & R100$\downarrow$ & MFN$\downarrow$ & Avg.$\downarrow$
& Ada.$\downarrow$ & QMag.$\downarrow$ & Elas.$\downarrow$ & R18$\downarrow$
& R50$\downarrow$ & R100$\downarrow$ & MFN$\downarrow$ & Avg.$\downarrow$ \\
\midrule
No Defense
& 95.39 & 89.40 & 95.85 & 94.24 & 96.08 & 95.85 & 93.78 & 94.37
& 76.98 & 30.16 & 76.98 & 72.75 & 82.54 & 85.71 & 61.11 & 69.46 \\

AdvDM
& \textbf{57.04} & 42.96 & \textbf{57.04} & \textbf{55.80}
& \textbf{57.78} & \textbf{57.78} & \underline{53.33} & \textbf{54.53}
& 63.86 & 19.84 & 69.02 & 59.78
& 73.64 & 77.17 & 43.75 & 58.15 \\

LowKey
& 95.97 & 88.63 & 96.45 & 95.02
& 96.45 & 95.97 & 93.13 & 94.52
& 76.47 & 27.81 & 77.01 & 70.05
& 79.14 & 82.89 & 52.94 & 66.62 \\

GIFT
& 72.62 & \textbf{12.86} & \underline{73.57} & 64.29
& 83.81 & 84.29 & 61.19 & 64.66
& \underline{28.57} & \underline{2.96} & \underline{29.11} & \underline{24.26}
& \textbf{32.61} & \textbf{38.54} & \underline{16.71} & \underline{24.68} \\

IDDM
& \underline{69.57} & \underline{21.12} & 73.64 & \underline{56.01}
& \underline{80.81} & \underline{83.33} & \textbf{48.64} & \underline{61.87}
& \textbf{25.91} & \textbf{1.55} & \textbf{29.02} & \textbf{19.69}
& \underline{35.23} & \underline{40.41} & \textbf{10.10} & \textbf{23.13} \\
\bottomrule
\end{tabular*}
\vspace{-4pt}
\end{table*}

\section{Experiments}
\label{sec:experiment} 

\begin{figure*}[t]
    \centering
    \includegraphics[width=0.98\linewidth]{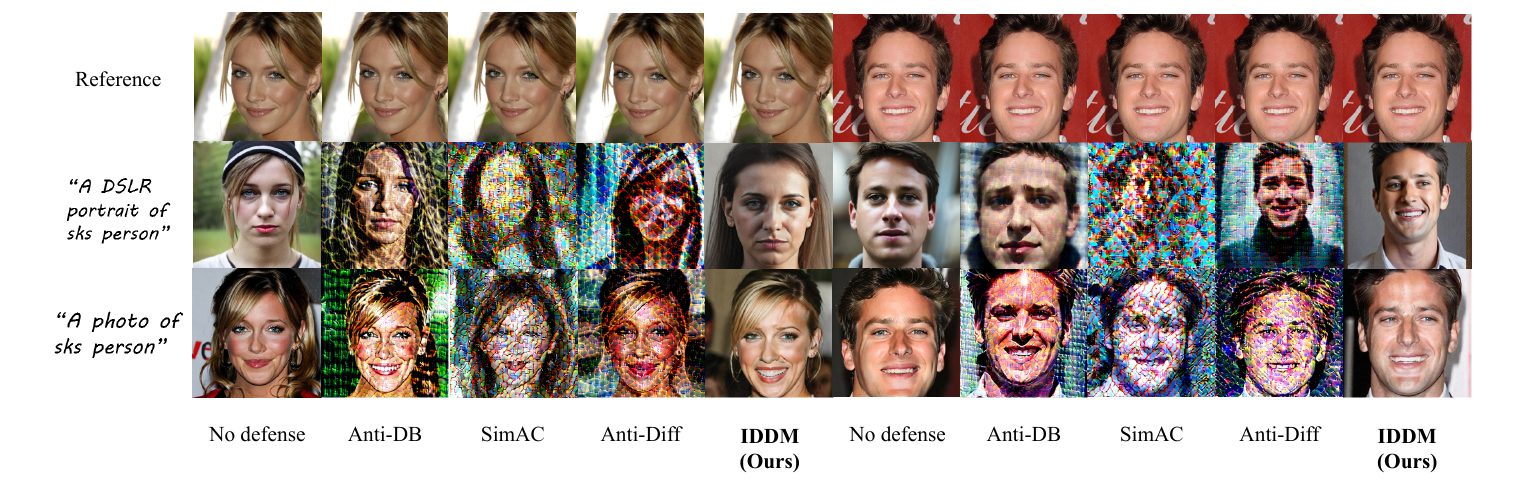}
    \caption{Qualitative defense results on CelebA-HQ. First row: reference image. Second row: generated images on the prompt ``a dslr portrait of sks person". Third row: generated images on the prompt ``a photo of sks person".}
    \label{cele_defense_results}
\end{figure*}
\subsection{Setup}
\noindent\textbf{Datasets.}
We consider two complementary facial datasets for experiments: CelebA-HQ \citep{karras2018progressive} and VGGFace2 \citep{cao2018vggface2}. CelebA-HQ provides high-quality face images for evaluating generation fidelity, while VGGFace2 is a large-scale in-the-wild face dataset with substantial intra-identity variation, making it suitable for evaluating identity linkability under more diverse pose and appearance conditions. For each dataset, we sample 50 identities and collect at least 16 face images per identity. For each identity, the images are partitioned into three disjoint subsets: (1) a fixed reference set $A$ (i.e., $\mathcal{X}_c$) of 4 images for short fine-tuning of the surrogate personalized diffusion model and computation of the identity prototype; (2) a protection set $B$ (i.e., $\mathcal{X}_0$) of 4 images for identity-decoupled data optimization; and (3) an additional clean set with at least 4 images for metric evaluation. All images are center-cropped and resized to $512\times512$.

\noindent\textbf{Baselines.}
To the best of our knowledge, existing defenses for personalized diffusion models mostly follow an \emph{anti-personalization} objective, i.e., preventing personalization from public reference images. In contrast, our setting is utility-preserving: personalization is still authorized for legitimate users, while identity traceability from generated outputs is reduced. As no prior defense is tailored to this setting, we compare against the closest available baselines from anti-personalization work, namely Anti-DreamBooth \citep{vanle2023antidreambooth}, SimAC \citep{wang2024simac}, and Anti-Diffusion \citep{zheng2025antidiffusion}. For brevity, we refer to Anti-DreamBooth as Anti-DB and Anti-Diffusion as Anti-Diff. Furthermore, we compare with some post-processing baselines on generations, including AdvDM~\citep{advdm}, LowKey~\citep{low}, and GIFT~\citep{gift}.

\noindent\textbf{Implementation details.}
We adopt Stable Diffusion v2.1-base as the pretrained backbone and DreamBooth \citep{ruiz2023dreambooth} for subject-driven personalization. Unless otherwise stated, we run $ K=10$ outer iterations with a stage split of $\rho=0.1$, and each outer iteration contains $8$ PGD update steps. For fair comparison, all methods use the same step size $\alpha=0.008$ and bounded budget $\eta=0.08$ at $512\times512$ resolution. 

For identity decoupling, we use four surrogate face recognizers: IRSE50 \citep{Hu_2018_CVPR}, IR152 \citep{He_2016_CVPR}, MobileFaceNet \citep{chen2018mobilefacenets}, and FaceNet \citep{schroff2015facenet}. Before feature extraction, we detect faces and 5-point landmarks using SCRFD \citep{guo2022scrfd}, align each face to a canonical template, and resize the aligned crop to match each input resolution required by each recognizer (e.g., $112\times112$ or $160\times160$). We set ensemble temperature $\tau=0.03$. For evaluation, we use various additional off-the-shelf face recognizers, including AdaFace \citep{kim2022adaface}, QMagFace \citep{Terhorst_2023_WACV}, ElasticFace \citep{Boutros_2022_CVPR}, R18, R50, and R100 \citep{zoo}, which are kept frozen and are strictly disjoint from the surrogate recognizers used during optimization.

The surrogate DreamBooth model is fine-tuned for 1000 steps with prior preservation (\ref{prior}) on the private image set, updating both the UNet and text encoder with batch size 2 and learning rate $5\times 10^{-7}$. The instance prompt is \textit{``a photo of [sks] person''}. All experiments are conducted on a single NVIDIA A100 GPU. Personalized diffusion model training takes about 20 minutes per identity, and inference takes about 1 minute per batch, making deployment practical for authorized users.

\noindent\textbf{Evaluation metrics.}
We evaluate both visual fidelity and identity linkability. For portrait utility, we report the face detection success rate (\textbf{FSR}), i.e., the fraction of generated images in which SCRFD \citep{guo2022scrfd} detects a face. For identity linkability, we use four metrics. \textbf{ISM} is the mean cosine similarity between each detected generated face and the reference embedding computed from the corresponding clean set using ArcFace \citep{deng2019arcface}. \textbf{ADA} is defined analogously using AdaFace \citep{kim2022adaface}. Lower ISM and ADA indicate weaker identity preservation and thus stronger privacy protection. We also report the top-$k$ untargeted identity retrieval success rate (\textbf{Rk-U}) with $k\in\{1,5\}$, which measures whether the correct identity appears among the top-$k$ retrieved candidates. Following \citep{tar1,tar2}, we conduct a threshold-based verification to access True Accept Rate \textbf{(TAR)} under practical FR decision thresholds. All identity-based metrics are computed on generated images with successful face detection.  
For visual fidelity, we report \textbf{FID} \citep{heusel2017ttur}, \textbf{SER-FIQ} \citep{terhorst2020serfiq}, and \textbf{BRISQUE} \citep{mittal2012brisque}, denoted as \textbf{SER} and \textbf{BRQ} for brevity. FID measures distribution-level fidelity, while SER-FIQ and BRISQUE assess perceptual and no-reference image quality, respectively.

\subsection{Experimental Results}
\label{r1}



\noindent\textbf{Main results.} We first compare IDDM with state-of-the-art anti-perso\allowbreak-nalization baselines across multiple metrics and datasets. Table~\ref{tab:defense_results} reports the main results on CelebA-HQ and VGGFace2 under the two primary prompts, and Table~\ref{tab:extra_prompts_rho} further evaluates IDDM under four additional prompts. Overall, IDDM achieves the most favorable balance between protection and generation quality.

Compared with standard DreamBooth (No Defense), all defended variants of IDDM consistently reduce identity leakage, as reflected by lower ISM and ADA across datasets and prompts. Compared with existing anti-personalization baselines, IDDM also preserves substantially stronger portrait usability and visual quality. Although aggressive baselines (i.e., SimAC and Anti-Diff) indeed achieve more extreme identity suppression, this comes at the cost of severely degraded usability and image quality. Anti-DB has moderate identity protection, yet still suffers from noticeable quality deterioration.

We further compare IDDM with post-processing privacy protection baselines in Table~\ref{tab:postprocess_main_metrics}. These are related to the model-side output immunization, as they directly process generated images after personalization.
The results in Table~\ref{tab:postprocess_main_metrics} indicate that IDDM achieves competitive performance, e.g., IDDM has the lowest ISM and ADA under \textit{``a dslr portrait of sks person"}. 

Moreover, $\rho$ provides an explicit and tunable handle over the privacy-utility operating point of IDDM. As shown in Table~\ref{tab:defense_results} and Table~\ref{tab:extra_prompts_rho}, smaller values of $\rho$ (e.g., $\rho=0.1$) generally yield stronger identity suppression, making them more suitable for high-privacy deployments, while larger values (e.g., $\rho=0.5$) tend to improve portrait usability and visual fidelity, making them preferable when generation quality is prioritized. The intermediate setting $\rho=0.3$ serves as a practical default, achieving reduced identity leakage over the DreamBooth personalization (without defense) while preserving strong portrait quality and usability. The complete sweep over $\rho \in \{0.1,0.2,0.3,0.5,0.7\}$ is provided in Appendix A.1. Overall, IDDM enables users to choose a deployment-oriented trade-off based on practical requirements. Qualitative examples in Fig.~\ref{cele_defense_results} further show that IDDM better preserves natural facial structure and appearance than competing defenses.

\noindent\textbf{Privacy protection analysis.}
We evaluate identity linkage risk by using retrieval-based and threshold-based face recognition protocols. The top-$k$ untargeted identity retrieval under multiple datasets and face recognizers are reported in Appendix A.2. In addition, Table~\ref{tab:new_evaluation} evaluates TAR at FAR=1e-4 across seven FR models. Here, FAR controls the false matching rate, i.e., how often different identities are incorrectly accepted as the same person, while TAR measures whether generated images are still accepted as the true identity under this decision threshold. The results show that our defense significantly reduces TAR compared with No Defense and achieves the best or second-best results among baselines, showing practical mitigation of identity linkage. 

\noindent\textbf{Personalization performance for unseen prompts.}
We further evaluate four unseen prompts beyond the training prompt, including hat, glasses, red hair, and big eyes. Table~\ref{tab:extra_prompts_rho} reports the quantitative results, and Figure~\ref{cele_6p} in Appendix~C shows qualitative examples on CelebA-HQ. Across all prompts, IDDM continues to generate diverse and visually plausible portraits while weakening identity linkability beyond the training prompt. 

\noindent\textbf{Protection robustness.}
We evaluate the robustness of generated images under common transformations, including JPEG compression, resizing, blurring, and cropping, with full results reported in Appendix A.2. IDDM achieves consistently low ADA and BRQ under these transformations, indicating robust identity protection.

\subsection{Ablation Study}
We study 1) the diffusion backbone, 2) several design choices and hyperparameters in IDDM, including the ablation of the diffusion backbone, the FR ensemble used for optimization, the stage split ratio $\rho$, the outer-loop iterations $K$, the bounded budget $\eta$, and the Softmin temperature $\tau$, and 3) the LoRA-based personalization method. Among them, $\rho$ is the primary control parameter for the privacy-utility trade-off and is discussed in the main text. The remaining ablations, together with detailed numerical results and discussions, are deferred to Appendix~A.1 and A.2.
\section{Conclusion}
\label{sec:conclusion}

In this paper, we study a new defense scenario in face-centered personalized diffusion and introduce \emph{model-side output immunization}. In contrast to anti-personalization defenses that block fine-tuning, our proposed IDDM enables authorized personalization while reducing downstream biometric linkability of the released outputs, with a tunable privacy-utility trade-off. IDDM alternates personalization updates with identity-decoupled data optimization under a two-stage schedule, yielding personalized outputs that remain visually useful while becoming less biometrically linkable. We evaluate the effectiveness of IDDM across datasets, prompts, and state-of-the-art face recognizers. Moreover, IDDM can be plugged into standard DreamBooth/LoRA pipelines to produce protected avatars by default, making privacy-compatible personalized creation the norm. We hope this work encourages broader adoption of utility-preserving privacy defenses in personalized diffusion systems and inspires future research on trustworthy generative personalization.

\clearpage
\bibliographystyle{ACM-Reference-Format}
\balance
\bibliography{main}

\clearpage
\appendix

\section{Additional Experimental Results}


\begin{table*}[t]
\centering
\caption{Defense performance of IDDM on CelebA-HQ under different values of the tunable ratio $\rho$. }
\label{iddm_ratio_metrics}
\setlength{\tabcolsep}{3.5pt}
\small
\begin{tabular}{l
S[table-format=1.2] S[table-format=1.2] S[table-format=1.2] S[table-format=3.2] S[table-format=1.2] S[table-format=2.2]
S[table-format=1.2] S[table-format=1.2] S[table-format=1.2] S[table-format=3.2] S[table-format=1.2] S[table-format=2.2]}
\toprule
\multirow{2}{*}{$\rho$} &
\multicolumn{6}{c}{\textit{``a photo of sks person''}} &
\multicolumn{6}{c}{\textit{``a dslr portrait of sks person''}} \\
\cmidrule(lr){2-7}\cmidrule(lr){8-13}
& {FSR$\uparrow$} & {ISM$\downarrow$} & {ADA$\downarrow$} & {FID$\downarrow$} & {SER$\uparrow$} & {BRQ$\downarrow$}
& {FSR$\uparrow$} & {ISM$\downarrow$} & {ADA$\downarrow$} & {FID$\downarrow$} & {SER$\uparrow$} & {BRQ$\downarrow$} \\
\midrule
No defense
& 0.57 & 0.69 & 0.66 & 158.09 & 0.76 & 10.65
& 0.50 & 0.48 & 0.44 & 214.35 & 0.74 & 4.88 \\

0.1
& {\textbf{0.65}} & {\textbf{0.43}} & {\textbf{0.38}} & 213.37 & 0.74 & {\textbf{16.81}}
& 0.48 & {\textbf{0.29}} & {\textbf{0.25}} & 237.54 & 0.75 & 11.71 \\

0.2
& 0.63 & 0.49 & 0.42 & 186.49 & 0.74 & 19.67
& 0.50 & 0.34 & 0.30 & 219.37 & 0.75 & {\textbf{11.19}} \\

0.3
& 0.62 & 0.51 & 0.47 & 172.80 & {\textbf{0.75}} & 21.31
& 0.52 & 0.36 & 0.33 & 208.40 & 0.75 & 12.56 \\

0.5
& 0.62 & 0.53 & 0.49 & 171.77 & {\textbf{0.75}} & 20.87
& {\textbf{0.56}} & 0.40 & 0.37 & 203.89 & 0.77 & 13.65 \\

0.7
& 0.59 & 0.55 & 0.51 & {\textbf{171.64}} & {\textbf{0.75}} & 20.49
& 0.54 & 0.45 & 0.41 & {\textbf{202.02}} & {\textbf{0.78}} & 14.99 \\
\bottomrule
\end{tabular}
\end{table*}

\begin{table*}
\centering
\caption{Effect of bounded budget on quality and identity-related metrics on CelebA-HQ.}
\label{budget_ablation}
\small
\setlength{\tabcolsep}{4pt}
\renewcommand{\arraystretch}{1.2}
\begin{threeparttable}
\begin{tabular}{@{}l
  S[table-format=2.2] S[table-format=1.2] S[table-format=1.2]
  S[table-format=1.2] S[table-format=1.2] S[table-format=1.2] S[table-format=3.2] S[table-format=2.2]
  S[table-format=1.2] S[table-format=1.2] S[table-format=1.2] S[table-format=3.2] S[table-format=2.2]
@{}}
\toprule
\multirow{2}{*}{$\eta$} &
\multicolumn{3}{c}{Quality} &
\multicolumn{5}{c}{\textit{``a photo of sks person''}} &
\multicolumn{5}{c}{\textit{``a dslr portrait of sks person''}} \\
\cmidrule(lr){2-4}\cmidrule(lr){5-9}\cmidrule(lr){10-14}
& {PSNR$\uparrow$} & {SSIM$\uparrow$} & {LPIPS$\downarrow$}
& {FSR$\uparrow$} & {ISM$\downarrow$} & {ADA$\downarrow$} & {FID$\downarrow$} & {BRQ$\downarrow$}
& {FSR$\uparrow$} & {ISM$\downarrow$} & {ADA$\downarrow$} & {FID$\downarrow$} & {BRQ$\downarrow$} \\
\midrule
0
& \multicolumn{1}{c}{--} & \multicolumn{1}{c}{--} & \multicolumn{1}{c}{--}
& 0.57 & 0.69 & 0.66 & 158.09 & 10.65
& 0.50 & 0.48 & 0.44 & 214.35 & 4.88 \\

0.05
& 35.12 & 0.90 & 0.11
& 0.60 & 0.49 & 0.43 & 203.15 & 21.78
& 0.46 & 0.37 & 0.33 & 217.82 & 7.17 \\

0.08\tnote{*}
& 32.83 & 0.84 & 0.21
& 0.65 & 0.43 & 0.38 & 213.37 & 16.81
& 0.48 & 0.29 & 0.25 & 237.54 & 11.71 \\
0.10
& 31.61 & 0.80 & 0.26
& 0.64 & 0.39 & 0.33 & 232.58 & 15.81
& 0.43 & 0.24 & 0.20 & 260.44 & 15.77 \\

0.15
& 30.00 & 0.74 & 0.34
& 0.70 & 0.31 & 0.26 & 248.72 & 15.83
& 0.44 & 0.19 & 0.15 & 281.81 & 19.63 \\
\bottomrule
\end{tabular}

\begin{tablenotes}[flushleft]
\footnotesize
\item[*] Default setting used in our main experiments.
\end{tablenotes}
\end{threeparttable}
\end{table*}

\begin{table*}[t]
\centering
\caption{Defense performance of IDDM on CelebA-HQ under different values of $\tau$ defined in Eq. (12). $^*$ denotes the default setting in our experiments.}\label{tau}
\label{tau}
\setlength{\tabcolsep}{3.5pt}
\small
\begin{tabular}{l
  S[table-format=1.2] S[table-format=1.2] S[table-format=1.2] S[table-format=3.2] S[table-format=1.2] S[table-format=2.2]
  S[table-format=1.2] S[table-format=1.2] S[table-format=1.2] S[table-format=3.2] S[table-format=1.2] S[table-format=2.2]}
\toprule
\multirow{2}{*}{$\tau$} &
\multicolumn{6}{c}{\textit{``a photo of sks person''}} &
\multicolumn{6}{c}{\textit{``a dslr portrait of sks person''}} \\
\cmidrule(lr){2-7}\cmidrule(lr){8-13}
& {FSR$\uparrow$} & {ISM$\downarrow$} & {ADA$\downarrow$} & {FID$\downarrow$} & {SER$\uparrow$} & {BRQ$\downarrow$}
& {FSR$\uparrow$} & {ISM$\downarrow$} & {ADA$\downarrow$} & {FID$\downarrow$} & {SER$\uparrow$} & {BRQ$\downarrow$} \\
\midrule
No defense & 0.57 & 0.69 & 0.66 & 158.09 & 0.76 & 10.65 & 0.50 & 0.48 & 0.44 & 214.35 & 0.74 &  4.88 \\
0.01      & 0.59 & 0.43 & 0.38 & 218.76 & 0.74 & 16.90 & \textbf{0.49} & 0.28 & 0.24 & 245.25 & 0.75 & 12.69 \\
0.03$^{*}$& \textbf{0.65} & 0.43 & 0.38 & \textbf{213.37} & 0.74 & 16.81 & 0.48 & 0.29 & 0.25 & \textbf{237.54} & 0.75 & \textbf{11.71} \\
0.05      & 0.64 & 0.43 & 0.38 & 213.52 & \textbf{0.75} & \textbf{16.70} & 0.48 & 0.27 & 0.23 & 239.04 & 0.75 & 12.18 \\
0.10      & 0.65 & \textbf{0.40} & \textbf{0.35} & 216.65 & 0.74 & 18.42 & 0.44 & \textbf{0.26} & \textbf{0.23} & 245.03 & 0.75 & 13.34 \\
\bottomrule
\end{tabular}
\end{table*}

\begin{table*}[t]
\centering
\caption{Ablation of the face recognition (FR) ensemble used during optimization. R50, R152, MF, and FN denote IRSE50, IR152, MobileFace, and FaceNet, respectively.}
\small
\setlength{\tabcolsep}{3pt}
\renewcommand{\arraystretch}{1.15}
\begin{tabular*}{\textwidth}{@{\extracolsep{\fill}}lcccccccccccc}
\toprule
\multirow{2}{*}{\makecell[l]{\textbf{Opt. FR}\\\textbf{ensemble}}}
& \multicolumn{6}{c}{\textit{``a photo of sks person''}}
& \multicolumn{6}{c}{\textit{``a dslr portrait of sks person''}} \\
\cmidrule(lr){2-7} \cmidrule(lr){8-13}
& FSR$\uparrow$ & ISM$\downarrow$ & ADA$\downarrow$ & FID$\downarrow$ & SER$\uparrow$ & BRQ$\downarrow$
& FSR$\uparrow$ & ISM$\downarrow$ & ADA$\downarrow$ & FID$\downarrow$ & SER$\uparrow$ & BRQ$\downarrow$ \\
\midrule
No Defense     
& 0.57 & 0.69 & 0.66 & 158.09 & 0.76 & 10.65
& 0.50 & 0.48 & 0.44 & 214.35 & 0.74 & 4.88 \\

R50            
& 0.66 & \textbf{0.35} & \textbf{0.30} & 228.07 & 0.73 & 18.90
& 0.42 & \textbf{0.23} & \textbf{0.19} & 249.17 & \textbf{0.76} & 14.15 \\

R50+R152       
& 0.64 & 0.38 & 0.33 & \textbf{212.31} & 0.72 & 19.19
& 0.44 & 0.24 & 0.20 & 247.83 & 0.74 & 13.08 \\

R50+R152+MF    
& 0.66 & 0.43 & 0.37 & 219.88 & 0.73 & 18.79
& 0.44 & 0.29 & 0.24 & 245.15 & 0.75 & 12.59 \\
R50+R152+MF+FN 
& 0.65 & 0.43 & 0.38 & 213.37 & \textbf{0.74} & 16.81
& \textbf{0.48} & 0.29 & 0.29 & \textbf{237.54} & 0.75 & \textbf{11.71} \\
\bottomrule
\end{tabular*}\label{fr_ablation}
\end{table*}

\begin{table}[t]
 \small
\setlength{\tabcolsep}{4.5pt}  
\renewcommand{\arraystretch}{1.2} 
\caption{Defense performance with different outer iterations on CelebA-HQ under prompt \emph{`` a photo of sks person"}.}\label{k_outer}
\begin{tabular}{c|cccccc}
\hline
$K$& FSR$\uparrow$  & ISM$\downarrow$& ADA$\downarrow$& FID$\downarrow$ & SER$\uparrow$  & BRQ$\downarrow$\\ \hline
No defense & 0.57 & 0.69 & 0.66 & 158.09 & 0.76 & 10.65 \\
10         & 0.65 & 0.43 & 0.38 & \textbf{213.37} & 0.74 & \textbf{16.81} \\
20         & 0.63 & 0.42 & 0.36 & 217.14 & 0.74 & 17.83 \\
30         & \textbf{0.67} & 0.38 & 0.33 & 228.20 & 0.72 & 18.59 \\
50         & 0.62 & 0.38 & 0.33 & 228.66 & 0.72 & 20.14     \\ \hline
\end{tabular}
\end{table}

\begin{table*}[t]
\centering
\caption{Defense performance comparison of IDDM across different diffusion model versions on CelebA-HQ.}
\label{version}
\small
\setlength{\tabcolsep}{1.8pt}
\renewcommand{\arraystretch}{1.0}
\begin{tabular}{@{}cccccccccccccc@{}}
\toprule
\multirow{2}{*}{\textbf{Version}} & \multirow{2}{*}{\textbf{Method}} &
\multicolumn{6}{c}{\textit{``a photo of sks person''}} &
\multicolumn{6}{c}{\textit{``a dslr portrait of sks person''}} \\
\cmidrule(lr){3-8} \cmidrule(lr){9-14}
& &
FSR$\uparrow$ & ISM$\downarrow$ & ADA$\downarrow$ & FID$\downarrow$ & SER$\uparrow$ & BRQ$\downarrow$ &
FSR$\uparrow$ & ISM$\downarrow$ & ADA$\downarrow$ & FID$\downarrow$ & SER$\uparrow$ & BRQ$\downarrow$ \\
\midrule

\multirow{2}{*}{V2.1}
& No Defense
& 0.57 & 0.69 & 0.66 & 158.09 & 0.76 & 10.65
& 0.50 & 0.48 & 0.44 & 214.35 & 0.74 & 4.88 \\
& Defense
& 0.65 & 0.43 & 0.38 & 213.37 & 0.74 & 16.81
& 0.48 & 0.29 & 0.25 & 237.54 & 0.75 & 11.71 \\

\midrule

\multirow{2}{*}{V1.5}
& No Defense
& 0.67 & 0.67 & 0.63 & 160.42 & 0.75 & 11.96
& 0.76 & 0.37 & 0.33 & 211.26 & 0.69 & 6.94 \\
&Defense
& 0.77 & 0.43 & 0.38 & 181.13 & 0.73 & 26.93
& 0.70 & 0.28 & 0.24 & 189.30 & 0.71 & 18.63 \\

\midrule

\multirow{2}{*}{V1.4}
& No Defense
& 0.69 & 0.66 & 0.63 & 160.53 & 0.75 & 13.56
& 0.69 & 0.34 & 0.30 & 225.13 & 0.67 & 7.99 \\
& Defense
& 0.75 & 0.45 & 0.41 & 181.90 & 0.74 & 23.17
& 0.67 & 0.29 & 0.25 & 199.84 & 0.69 & 19.20 \\
\bottomrule
\end{tabular}
\end{table*}

\begin{table*}[t]
\caption{Comparing the defense performance between the joint optimization and the stage-wise optimization on CelebA-HQ under a convenient setting.}\label{reason}
\centering
\setlength{\tabcolsep}{3pt}        
\renewcommand{\arraystretch}{1.15} 

\begin{tabular}{c|cccccc|cccccc}
\hline
\multirow{2}{*}{Method} &
\multicolumn{6}{c|}{\textit{``a photo of sks person"}} &
\multicolumn{6}{c}{\textit{``a dslr portrait of sks person"}} \\ \cline{2-13}
& FSR$\uparrow$ & ISM$\downarrow$ & ADA$\downarrow$ & FID$\downarrow$ & SER$\uparrow$ & BRQ$\downarrow$
& FSR$\uparrow$ & ISM$\downarrow$ & ADA$\downarrow$ & FID$\downarrow$ & SER$\uparrow$ & BRQ$\downarrow$ \\ \hline

No defense      & 0.57 & 0.69 & 0.66 & 158.09 & 0.76 & 10.65 & 0.50 & 0.48 & 0.44 & 214.35 & 0.74 & 4.88 \\
Joint Opt.      & 0.65 & 0.43 & 0.39 & 220.14 & \textbf{0.75}    & 16.97     & 0.38 & 0.29 & 0.25 & 250.06 & 0.75    & 13.27     \\

\rowcolor{gray!15}
Stage-wise Opt. & \textbf{0.65} & \textbf{0.43} & \textbf{0.38} & \textbf{213.37} & 0.74& \textbf{16.81}
               & \textbf{0.48} & \textbf{0.29} & \textbf{0.25} & \textbf{237.54} & \textbf{0.75} & \textbf{11.71} \\ \hline

\end{tabular}
\end{table*}
\begin{table*}[t]
\centering
\caption{Defense performance of IDDM under LoRA-based personalization with different values of the tunable ratio $\rho$ on CelebA-HQ.}
\label{lora}
\setlength{\tabcolsep}{2pt}
\renewcommand{\arraystretch}{1.2}
\small
\begin{tabular*}{\textwidth}{@{\extracolsep{\fill}}l*{12}{c}@{}}
\toprule
\multirow{2}{*}{\textbf{Method}}
& \multicolumn{6}{c}{\textit{``a photo of sks person''}} 
& \multicolumn{6}{c}{\textit{``a dslr portrait of sks person''}} \\
\cmidrule(lr){2-7} \cmidrule(lr){8-13}
& FSR$\uparrow$ & ISM$\downarrow$ & ADA$\downarrow$ & FID$\downarrow$ & SER$\uparrow$ & BRQ$\downarrow$
& FSR$\uparrow$ & ISM$\downarrow$ & ADA$\downarrow$ & FID$\downarrow$ & SER$\uparrow$ & BRQ$\downarrow$ \\
\midrule
No Defense
& 0.59 & 0.62 & 0.58 & 179.41 & 0.77 & 12.60
& 0.70 & 0.16 & 0.13 & 266.44 & 0.70 & 10.86 \\

IDDM ($\rho{=}0.1$)
& \textbf{0.68} & \textbf{0.43} & \textbf{0.38} & 236.83 & 0.76 & 19.19
& 0.69 & \textbf{0.10} & \textbf{0.08} & 279.48 & 0.69 & 18.03 \\

IDDM ($\rho{=}0.3$)
& 0.59 & 0.55 & 0.51 & 164.25 & \textbf{0.79} & 18.98
& \textbf{0.70} & 0.12 & 0.10 & 259.69 & \textbf{0.72} & 13.81 \\

IDDM ($\rho{=}0.5$)
& 0.58 & 0.59 & 0.56 & \textbf{160.16} & 0.78 & \textbf{17.19}
& 0.69 & 0.14 & 0.12 & \textbf{250.34} & 0.71 & \textbf{12.54} \\
\bottomrule
\end{tabular*}
\end{table*}


\begin{table*}[t]
\centering
\caption{Top-$k$ untargeted identity retrieval attack success rates (\%, $k{=}1,5$) on CelebA-HQ across AdaFace \citep{kim2022adaface}, QMagFace \citep{Terhorst_2023_WACV}, and ElasticFace \cite{Boutros_2022_CVPR}. The gallery contains 112 identities, including the evaluated identities. Lower is better; best results are in \textbf{bold} and second-best are \underline{underlined}.}
\label{tab:topk_adaptive}
\setlength{\tabcolsep}{1.5pt}
\renewcommand{\arraystretch}{1.2}
\small
\begin{tabular*}{\textwidth}{@{\extracolsep{\fill}}lcccccccccccccccc}
\toprule
\multirow{3}{*}{\textbf{Method}} 
& \multicolumn{8}{c}{\textit{``a photo of sks person''}}
& \multicolumn{8}{c}{\textit{``a dslr portrait of sks person''}} \\
\cmidrule(lr){2-9} \cmidrule(lr){10-17}
& \multicolumn{2}{c}{\textbf{AdaFace}} 
& \multicolumn{2}{c}{\textbf{QMagFace}} 
& \multicolumn{2}{c}{\textbf{ElasticFace}} 
& \multicolumn{2}{c}{\textbf{Average}}
& \multicolumn{2}{c}{\textbf{AdaFace}} 
& \multicolumn{2}{c}{\textbf{QMagFace}} 
& \multicolumn{2}{c}{\textbf{ElasticFace}} 
& \multicolumn{2}{c}{\textbf{Average}} \\
& R1-U & R5-U & R1-U & R5-U & R1-U & R5-U & R1-U & R5-U
& R1-U & R5-U & R1-U & R5-U & R1-U & R5-U & R1-U & R5-U \\
\midrule
No Defense 
& 87.06 & 87.06 & 87.06 & 88.24 & 89.41 & 89.41 & 87.84 & 88.24
& 88.61 & 89.87 & 89.87 & 96.20 & 94.94 & 96.20 & 91.14 & 94.09 \\

Anti-DB
& 77.78 & 88.89 & 88.89 & 88.89 & 88.89 & 88.89 & 85.19 & 88.89
& 80.56 & \underline{83.33} & 77.78 & 91.67 & 80.56 & 91.67 & 79.63 & 88.89 \\

SimAC
& 83.33 & \underline{83.33} & 83.33 & \underline{83.33} & 66.67 & 100.00 & 77.78 & 88.89
& \textbf{0} & \textbf{0} & \textbf{0} & \textbf{0} & \textbf{0} & \textbf{0} & \textbf{0} & \textbf{0} \\

Anti-Diff
& \textbf{60.00} & \textbf{60.00} & \textbf{60.00} & \textbf{60.00}
& \underline{60.00} & \textbf{60.00} & \textbf{60.00} & \textbf{60.00}
& 75.00 & 100.00 & 75.00 & \underline{75.00} & 75.00 & 100.00 & 75.00 & 91.67 \\

Ours
& \underline{67.53} & 84.42 & \underline{62.34} & 87.01
& \textbf{58.44} & \underline{88.31} & \underline{62.77} & \underline{86.58}
& \underline{67.53} & 84.42 & \underline{62.34} & 87.01 & \underline{58.44} & \underline{88.31} & \underline{62.77} & \underline{86.58} \\
\bottomrule
\end{tabular*}
\label{cele_topk_new}
\end{table*}

\begin{table*}[t]
\centering
\caption{Top-$k$ untargeted identity retrieval attack success rates (\%, $k{=}1,5$) on VGGFace2 across AdaFace \citep{kim2022adaface}, QMagFace \citep{Terhorst_2023_WACV}, and ElasticFace \cite{Boutros_2022_CVPR}. Lower is better; best results are in \textbf{bold} and second-best are \underline{underlined}.}
\label{tab:vgg_topk_new}
\setlength{\tabcolsep}{1.5pt}
\renewcommand{\arraystretch}{1.2}
\small
\begin{tabular*}{\textwidth}{@{\extracolsep{\fill}}lcccccccccccccccc}
\toprule
\multirow{3}{*}{\textbf{Method}}
& \multicolumn{8}{c}{\textit{`a photo of sks person''}}
& \multicolumn{8}{c}{\textit{`a dslr portrait of sks person''}} \\
\cmidrule(lr){2-9} \cmidrule(lr){10-17}
& \multicolumn{2}{c}{\textbf{AdaFace}}
& \multicolumn{2}{c}{\textbf{QMagFace}}
& \multicolumn{2}{c}{\textbf{ElasticFace}}
& \multicolumn{2}{c}{\textbf{Average}}
& \multicolumn{2}{c}{\textbf{AdaFace}}
& \multicolumn{2}{c}{\textbf{QMagFace}}
& \multicolumn{2}{c}{\textbf{ElasticFace}}
& \multicolumn{2}{c}{\textbf{Average}} \\
& R1-U & R5-U & R1-U & R5-U & R1-U & R5-U & R1-U & R5-U
& R1-U & R5-U & R1-U & R5-U & R1-U & R5-U & R1-U & R5-U \\
\midrule
No Defense
& 99.33 & 99.78 & 99.56 & 99.78 & 99.10 & 99.33 & 99.33 & 99.63
& 93.98 & 95.58 & 94.38 & 97.59 & 93.17 & 95.98 & 93.84 & 96.38 \\

Anti-DB
& \underline{96.62} & \textbf{97.30} & \textbf{96.62} & \textbf{97.97} & \textbf{95.95} & \underline{97.30}
& \textbf{96.40} & \underline{97.52}
& 83.72 & 90.70 & 86.82 & 90.70 & 85.27 & 92.25 & 85.27 & 91.22 \\

SimAC
& 98.04 & \underline{98.04} & 98.04 & \underline{98.04} & \underline{96.08} & \textbf{96.08}
& \underline{97.39} & \textbf{97.39}
& \textbf{0} & \textbf{33.33} & \textbf{0} & \textbf{33.33} & \textbf{0} & \textbf{0} & \textbf{0} & \textbf{22.22} \\

Anti-Diff
& \textbf{95.83} & 100.00 & 100.00 & 100.00 & 97.92 & 100.00
& 97.92 & 100.00
& \underline{78.33} & 90.00 & 80.00 & \underline{85.00} & \underline{70.00} & \underline{81.67} & 76.11 & \underline{85.56} \\

Ours
& 96.96 & 99.07 & \underline{97.67} & 99.30 & 97.67 & 99.30
& 97.43 & 99.22
& 78.52 & \underline{88.89} & \underline{75.56} & 90.37 & 72.22 & 87.41 & \underline{75.43} & 88.89 \\
\bottomrule
\end{tabular*}
\end{table*}

\begin{table*}[t]
\centering
\caption{Top-$k$ untargeted identity retrieval attack success rates (\%, $k{=}1,5$) on CelebA-HQ across four face recognition models. The gallery contains 112 identities including the evaluated identities. Lower is better; best results are in \textbf{bold} and second-best are \underline{underlined}.}
\label{tab:topk_celebahq}
\setlength{\tabcolsep}{1.5pt}
\renewcommand{\arraystretch}{1.2}
\small
\begin{tabular*}{\textwidth}{@{\extracolsep{\fill}}lcccccccccc}
\toprule
\multirow{2}{*}{\textbf{Method}} & \multicolumn{10}{c}{\textit{``a photo of sks person''}}\\
\cmidrule(lr){2-11}
& \multicolumn{2}{c}{\textbf{IRSE50}} & \multicolumn{2}{c}{\textbf{IR152}} & \multicolumn{2}{c}{\textbf{FaceNet}} & \multicolumn{2}{c}{\textbf{MobileFace}} & \multicolumn{2}{c}{\textbf{Average}}\\
& R1-U & R5-U & R1-U & R5-U & R1-U & R5-U & R1-U & R5-U & R1-U & R5-U\\
\midrule
No Defense & 100.00 & 100.00 & 100.00 & 100.00 & 98.00 & 100.00 & 99.50 & 100.00 & 99.38 & 100.00 \\
Anti-DB   & 100.00 & 100.00 & 96.43 & 100.00 & 92.86 & 92.86 & 92.86 & 100.00 & 95.54 & 98.22 \\
SimAC      & \underline{94.12} & \underline{94.12} & \textbf{76.47} & \textbf{88.24} & \underline{70.59} & \underline{82.35} & \underline{82.35} & \underline{94.12} & \underline{80.88} & \underline{89.71} \\
Anti-Diff  & 100.00 & 100.00 & \underline{80.00} & 100.00 & 80.00 & 90.00 & 100.00 & 100.00 & 90.00 & 97.50 \\
Ours       & \textbf{85.35} & \textbf{90.45} & 84.71 & \underline{93.63} & \textbf{54.14} & \textbf{77.07} & \textbf{75.80} & \textbf{90.45} & \textbf{75.00} & \textbf{87.90} \\
\midrule

\multirow{2}{*}{\textbf{Method}} & \multicolumn{10}{c}{\textit{``a dslr portrait of sks person''}}\\
\cmidrule(lr){2-11}
& \multicolumn{2}{c}{\textbf{IRSE50}} & \multicolumn{2}{c}{\textbf{IR152}} & \multicolumn{2}{c}{\textbf{FaceNet}} &
  \multicolumn{2}{c}{\textbf{MobileFace}} & \multicolumn{2}{c}{Average}\\
& R1-U & R5-U & R1-U & R5-U & R1-U & R5-U & R1-U & R5-U & R1-U & R5-U\\
\midrule
No Defense & 93.02 & 97.29 & 89.15 & 95.74 & 82.56 & 92.64 & 90.70 & 95.74 & 88.86 & 95.35 \\
Anti-DB    & 94.78 & 98.51 & 82.84 & 93.28 & 66.42 & 83.58 & 85.07 & 96.27 & 82.28 & 92.91 \\
SimAC      & \underline{71.43} & \textbf{85.71} & \textbf{42.86} & \textbf{71.43} & \textbf{28.57} & \textbf{42.86} & \textbf{57.14} & \textbf{57.14} & \textbf{50.00} & \textbf{64.29} \\
Anti-Diff  & 81.25 & 93.75 & 56.25 & \underline{75.00} & 43.75 & 62.50 & 65.63 & 81.25 & 61.72 & 78.13 \\
Ours       & \textbf{70.67} & \underline{89.42} & \underline{54.81} & 77.40 & \underline{35.10} & \underline{61.54} & \underline{58.17} & \underline{80.77} & \underline{54.69} &  \underline{77.28} \\
\bottomrule
\end{tabular*}
\end{table*}


\begin{table*}[t]
\centering
\caption{Top-$k$ untargeted identity retrieval attack success rates (\%, $k{=}1,5$) on VGGFace2 across four face recognition models. The gallery contains 100 identities (including evaluated identities). Lower is better; best results are in \textbf{bold} and second-best are \underline{underlined}.}
\label{tab:topk_vggface2}
\setlength{\tabcolsep}{1.5pt}
\renewcommand{\arraystretch}{1.2}
\small

\begin{tabular*}{\textwidth}{@{\extracolsep{\fill}}lcccccccccc}
\toprule
\multirow{2}{*}{\textbf{Method}} & \multicolumn{10}{c}{\textit{``a photo of sks person''}}\\
\cmidrule(lr){2-11}
& \multicolumn{2}{c}{\textbf{IRSE50}} & \multicolumn{2}{c}{\textbf{IR152}} & \multicolumn{2}{c}{\textbf{FaceNet}} & \multicolumn{2}{c}{\textbf{MobileFace}} & \multicolumn{2}{c}{\textbf{Average}}\\
& R1-U & R5-U & R1-U & R5-U & R1-U & R5-U & R1-U & R5-U & R1-U & R5-U\\
\midrule
No Defense & 99.78 & 99.78 & 98.88 & 99.33 & 97.54 & 98.88 & 99.10 & 99.78 & 98.83 & 99.44 \\
Anti-DB    & \underline{97.30} & \textbf{97.97} & 95.27 & \underline{95.95} & 90.54 & 95.95 & 95.27 & 98.65 & 94.60 & 97.13 \\
SimAC      & \textbf{96.08} & \underline{98.04} & \textbf{92.16} & \textbf{92.16} & \textbf{86.27} & \underline{92.16} & 94.12 & \textbf{96.08} & \textbf{92.16} & \underline{94.61} \\
Anti-Diff  & 97.92 & 100.00 & 93.75 & 97.92 & \underline{87.50} & 93.75 & \textbf{89.58} & 97.92 & \underline{92.19} & 97.40 \\
Ours       & 98.14 & 99.30 & \underline{93.24} & 97.90 & 97.53 & \textbf{80.89} & \underline{91.14} & \underline{97.44} & 95.01 & \textbf{93.88} \\
\midrule
\multirow{2}{*}{\textbf{Method}} & \multicolumn{10}{c}{\textit{``a dslr portrait of sks person''}}\\
\cmidrule(lr){2-11}
& \multicolumn{2}{c}{\textbf{IRSE50}} & \multicolumn{2}{c}{\textbf{IR152}} & \multicolumn{2}{c}{\textbf{FaceNet}} &
  \multicolumn{2}{c}{\textbf{MobileFace}} & \multicolumn{2}{c}{\textbf{Average}}\\
& R1-U & R5-U & R1-U & R5-U & R1-U & R5-U & R1-U & R5-U & R1-U & R5-U\\
\midrule
No Defense & 93.98 & 97.19 & 90.36 & 93.98 & 83.94 & 93.98 & 90.76 & 97.19 & 89.76 & 95.59 \\
Anti-DB    & 83.72 & 91.47 & 82.17 & 87.60 & 69.77 & 82.95 & 79.07 & 90.70 & 78.68 & 88.18 \\
SimAC      & \textbf{0.00} & \textbf{33.33} & \textbf{0.00} & \textbf{33.33} & \textbf{0.00} & \textbf{66.67} & \textbf{0.00} & \textbf{66.67} & \textbf{0.00} & \textbf{50.00} \\
Anti-Diff  & \underline{73.33} & \underline{88.33} & 73.33 & 85.00 & 65.00 & \underline{80.00} & 71.67 & \underline{83.33} & 70.83 & \underline{84.17} \\
Ours       & 81.11 & 94.07 & \underline{67.41} & \underline{83.70} & \underline{57.41} & 81.48 & \underline{71.48} & 91.11 & \underline{69.35} & 87.59 \\
\bottomrule
\end{tabular*}
\end{table*}

\begin{table}[!htbp]
\centering
\caption{Defense performance comparison with post-protection baselines under various transformations. ADA denotes the similarity scores on AdaFace, BRQ denotes the visual quality metric, and No Preproc. denotes results without transformations.}
\label{robustness}
\normalsize
\setlength{\tabcolsep}{1pt}
\renewcommand{\arraystretch}{1.2}
\resizebox{\linewidth}{!}{%
\begin{tabular}{@{}l|>{\columncolor{gray!12}}c cccc|>{\columncolor{gray!12}}c cccc@{}}
\toprule
\multirow{2}{*}{Method}
& \multicolumn{5}{c|}{ADA$\downarrow$}
& \multicolumn{5}{c}{BRQ$\downarrow$} \\
\cmidrule(lr){2-6} \cmidrule(lr){7-11}
& No Preproc. & JPEG & Resize & Blur & Crop
& No Preproc. & JPEG & Resize & Blur & Crop \\
\midrule
No Defense
& $0.44$ & \multicolumn{4}{c|}{--}
& $4.88$ & \multicolumn{4}{c}{--} \\

AdvDM
& $0.39$ & $0.38$ & $0.36$ & $0.38$ & $0.39$
& $14.82$ & $33.14$ & $16.18$ & $23.21$ & $64.23$ \\

LowKey
& $0.42$ & $0.39$ & $0.40$ & $0.33$ & $0.41$
& $17.14$ & $\mathbf{17.51}$ & $18.98$ & $23.32$ & $58.85$ \\

GIFT
& $0.26$ & $0.26$ & $0.25$ & $0.26$ & $0.26$
& $39.62$ & $55.27$ & $41.41$ & $40.15$ & $64.84$ \\

IDDM
& $\mathbf{0.25}$ & $\mathbf{0.25}$ & $\mathbf{0.23}$ & $\mathbf{0.25}$ & $\mathbf{0.25}$
& $\mathbf{11.71}$ & $26.21$ & $\mathbf{12.88}$ & $\mathbf{17.07}$ & $\mathbf{58.43}$ \\
\bottomrule
\end{tabular}%
}
\end{table}


\subsection{Ablation Study}
We provide additional ablations on four remaining design factors: the tunable stage ratio $\rho$ controlling privacy-utility trade-off, the bounded budget $\eta$ for internal data optimization, the softmin temperature $\tau$ for ensemble identity aggregation, the face recognition (FR) ensemble used during optimization, and the number of outer iterations $K$ in the alternating optimization loop. These results complement the main analysis by examining the sensitivity, effectiveness, and stability of IDDM under different design choices.

\noindent\textbf{Tunable stage ratio $\rho$.}
We provide the complete results on different $\rho$ values in Table~\ref{iddm_ratio_metrics}, forming a denser sweep over eight PGD steps. 
Larger $\rho$ weakens the protection
by assigning more fidelity-preserving steps (Stage I) and less identity-decoupling steps (Stage II), allowing IDDM
to recover more global face structure and identity-related information,
as reflected by improved FID and SER. This may alter local texture
statistics measured by BRQ, leading to worse BRQ scores.

\noindent\textbf{Bounded budgets $\eta$.}
We vary the $\ell_\infty$ budget $\eta$ to study its effect on the optimized image set $\mathcal{X}'$ and the resulting personalized model (Table~\ref{budget_ablation}). Here, $\eta$ determines the feasible range of the \emph{internal} data optimization in our personalization pipeline. Since the optimized set $\mathcal{X}'$ is never exposed to the attacker and is only used internally to introduce identity-decoupled supervision, PSNR/SSIM/LPIPS mainly reflect how far $\mathcal{X}'$ deviates from the original references. As $\eta$ increases, identity-related cues are generally suppressed more strongly, but excessively large budgets may also harm fidelity and generation quality. This ablation verifies the impact of the optimization budget on privacy preservation and visual quality. We use $\eta=0.08$ as the default since it provides a favorable balance in practice.

\noindent\textbf{Softmin temperature $\tau$.}
We ablate the softmin temperature $\tau$ used in the ensemble identity loss (Table~\ref{tau}). A smaller $\tau$ places more emphasis on the most identity-sensitive face recognizer, while a larger $\tau$ produces smoother weights across the ensemble. The results are relatively stable for $\tau \in [0.01, 0.10]$, suggesting that IDDM is not overly sensitive to this parameter. We set $\tau=0.03$ by default as it consistently provides strong identity suppression with competitive image quality.

\noindent\textbf{FR ensemble for optimization.}
We further ablate the FR ensemble used in the identity-decoupling optimization (Table~\ref{fr_ablation}). Specifically, we progressively expand the ensemble from a single recognizer (R50) to a four-model ensemble (R50+R152+MF+FN), where R50, R152, MF, and FN denote IRSE50, IR152, MobileFace, and FaceNet, respectively. Using only R50 yields the strongest identity suppression on several metrics, suggesting that a single strong recognizer already provides effective identity guidance. However, expanding the ensemble generally improves the overall balance across utility and visual quality metrics. In particular, the full four-model ensemble achieves the most balanced overall performance while maintaining competitive identity suppression. This suggests that combining diverse FR backbones makes the optimization target more robust, and we adopt the full ensemble by default.

\begin{algorithm}[b]
\caption{Alternating Optimization for IDDM}
\label{outer_k}
\begin{algorithmic}[1]
\Input Protection set $\mathcal{X}_0$; reference set $\mathcal{X}_c$;
pretrained diffusion model $M_{\Theta_0}$; outer iterations $K$; noise budget $\eta$; step size $\alpha$; total PGD steps $T$;
tunable stage ratio $\rho\in(0,1)$.
\Output trained personalized model $M_{\Theta_K}$.

\State $T_{\mathrm{rec}} \gets \lfloor \rho T \rfloor$
\For{$k=1$ \textbf{to} $K$}
  \State \textit{Short fine-tuning:}
  \State \hspace{1.2em}$\Theta'_k \gets \operatorname{FT}(\Theta_{k-1};\mathcal{X}_c)$ 
  \State \textit{Identity-Decoupled Data Optimization:}
  \State \hspace{1.2em}$\mathcal{X}' \gets \mathcal{X}_0$
  \For{$t=1$ \textbf{to} $T$}
    \If{$t \le T_{\mathrm{rec}}$}
      \State \hspace{1.2em}$\mathcal{X}' \gets \Pi_{\eta}\!\Big(\mathcal{X}' - \alpha\,\mathrm{sign}(\nabla_{\mathcal{X}'} \mathcal{L}_{\mathrm{rec}}(\Theta'_k,\mathcal{X}'))\Big)$
    \Else
      \State \hspace{1.2em}$\mathcal{X}' \gets \Pi_{\eta}\!\Big(\mathcal{X}' - \alpha\,\mathrm{sign}(\nabla_{\mathcal{X}'} \mathcal{L}_{\mathrm{id}}(\Theta'_k,\mathcal{X}'))\Big)$
    \EndIf
  \EndFor
  \State \textit{Update model:}
  \State \hspace{1.2em}$\Theta_k \gets \operatorname{FT}(\Theta'_k;\mathcal{X}')$
  \State $\mathcal{X}_0 \gets \mathcal{X}'$
\EndFor
\State \Return $M_{\Theta_K}$
\end{algorithmic}
\end{algorithm}

\noindent\textbf{Outer iterations $K$.}
We further study the number of outer-loop iterations $K$ during the alternating loop (Table~\ref{k_outer}). This parameter controls how long the optimized set $\mathcal{X}'$ is updated to track the changing model parameters during personalization. Increasing $K$ generally improves the robustness of the learned identity-decoupled supervision, but also increases computation and may gradually reduce fidelity. The gains diminish once optimization approaches convergence. We therefore adopt $K=10$ as a moderate default that balances protection, quality, and efficiency.

\noindent\textbf{Personalized diffusion backbone version.} Since Stable Diffusion is a powerful open-source text-to-image diffusion model, it is a representative choice for studying realistic personalization defenses. To evaluate effectiveness across model versions, we test IDDM on three widely used Stable Diffusion backbones, including v2.1, v1.5, and v1.4. Table \ref{version} shows that IDDM consistently reduces identity-related scores (i.e., ISM and ADA) compared to no-defense setting. At the same time, the generation quality remains competitive across different model versions under both prompts. It is concluded that our defense is not tied to a specific diffusion version and transfers well across different pretrained diffusion models and their standard variants.

\subsection{Other Results}
We provide additional results on the optimization strategy, LoRA-based personalization method, identity retrieval, and robustness of generations. Table~\ref{reason} compares joint optimization and stage-wise optimization, showing that the stage-wise design achieves consistently better or comparable results under both prompts, with clearer gains in FID and BRQ, indicating a more stable optimization process in our setting. Table \ref{lora} shows that IDDM remains effective under LoRA-based personalization. Table~\ref{cele_topk_new} and ~\ref{tab:vgg_topk_new} further report Top-$k$ untargeted identity retrieval on three face recognition models that are not used in optimization, where IDDM remains competitive and shows non-trivial transferability, especially under \textit{``a dslr portrait of sks person''}. Tables~\ref{tab:topk_celebahq} and~\ref{tab:topk_vggface2} report retrieval results on the four face recognition models used in optimization, showing that IDDM effectively reduces identity retrieval success on both CelebA-HQ and VGGFace2, while remaining competitive with prior defenses. Table~\ref{robustness} show that generations of IDDM remain robust under various transformations and achieve strong defense performance (i.e., low ADA and BRQ), indicating effective identity protection with better visual quality.

\section{Implementation of IDDM}
Algorithm~\ref{outer_k} summarizes the alternating optimization procedure of IDDM, which alternates between short personalization fine-tuning and identity-decoupled data optimization.

\section{Additional Qualitative Results}
\begin{figure*}
    \centering
    \includegraphics[width=1\linewidth]{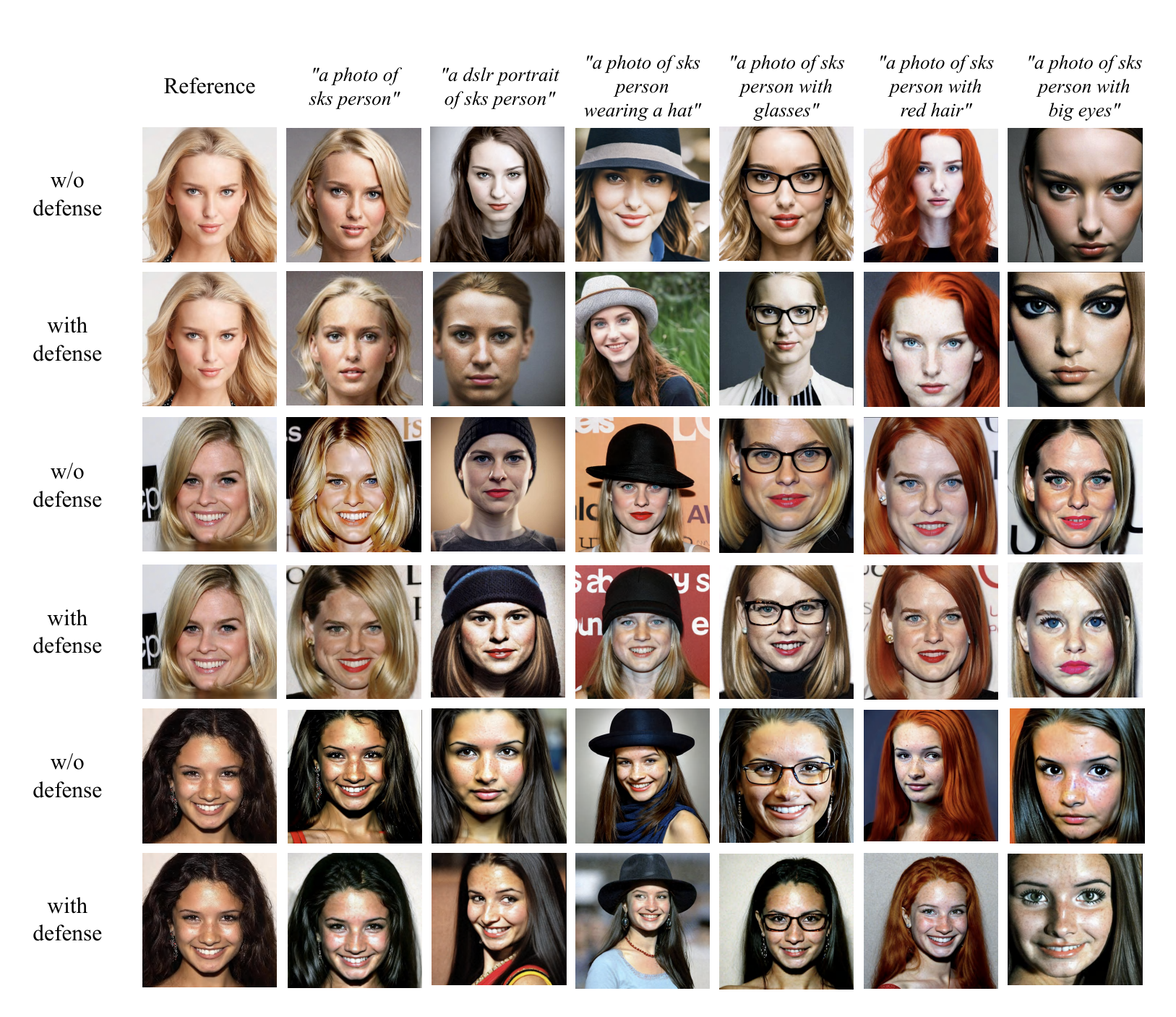}
    \caption{Comparison of generated images without defense and with defense under different prompts on CelebA-HQ.}
    \label{cele_6p}
\end{figure*}

Figure~\ref{cele_6p} compares the generated images without defense and with defense under six prompts on CelebA-HQ. The first prompt, \textit{``a photo of sks person''}, is the instance prompt used for training, while the other five prompts are additional prompts used only at inference time. For prompts such as \textit{``wearing a hat''}, \textit{``with glasses''}, and \textit{``with red hair''}, the defended results largely preserve the main visual characteristics of the generated subject, since these prompts mainly modify attributes or accessories rather than core facial features. For \textit{``with big eyes''}, the visual change is naturally more noticeable since the prompt directly alters facial features. Even in this case, the defended result does not appear to shift toward another specific identity, which is consistent with our untargeted setting: the goal is to suppress identity leakage rather than redirect the generation to a target identity.

\section{Ethical Considerations}

This work studies a new privacy-oriented defense for \emph{authorized personalization}. IDDM aims to reduce the identity linkability of public generations while preserving the utility of personalized generation, with tunable control over the privacy-utility balance to meet users' needs.

\noindent\textbf{Potential benefits and risks.}
The method may help legitimate users share personalized avatars with lower risk of identity linkage, tracking, or profiling. However, it is also potentially dual-use and could be misused to make generated faces harder to identify in harmful settings. Moreover, IDDM does not guarantee anonymity, since stronger or adaptive face recognition systems may still recover identity cues.

\noindent\textbf{Experimental scope and responsible release.}
All experiments are conducted on public datasets (CelebA-HQ and VGGFace2) in a controlled setting. We do not collect human-subject data or deploy the method in real-world surveillance scenarios. We scope IDDM explicitly to authorized personalization, do not claim complete anonymity, and do not release identity-specific protected images or personalized checkpoints tied to real individuals.

\section{Discussion and Future Work}

A natural alternative is to protect personalized generations with face recognition adversarial examples or other post-processing perturbations. Such methods typically operate by adding image-level perturbations before release. As a result, they require every public image to be protected consistently, and their effectiveness may be weakened by subsequent denoising, purification, or editing pipelines. In contrast, IDDM is a model-side defense: it directly produces personalized generations with reduced identity linkability, without relying on an additional perturbation step after generation. This avoids making protection dependent on whether the perturbation can survive later processing.

More broadly, we hope this work can encourage further study of privacy protection in personalized generation. Future research may explore more general model-side defenses, stronger evaluation under diverse generation and post-processing pipelines, and more principled ways to balance privacy and generation utility.


\end{document}